\documentclass[10pt,journal,compsoc]{IEEEtran}
\usepackage{array}
\usepackage{mdwmath}
\usepackage{mdwtab}
\usepackage{eqparbox}
\usepackage{makecell}

\ifCLASSOPTIONcompsoc
  \usepackage[nocompress]{cite}
\else
  \usepackage{cite}
\fi
\ifCLASSINFOpdf
   \usepackage[pdftex]{graphicx}
\else
\fi
\usepackage{amsfonts}
\usepackage{bbm}

\usepackage{amsmath}
\begin{document}
%
\title{Attribute And-Or Grammar for Joint Parsing of Human Attributes, Part and Pose}
%
%
%
%

\author{Seyoung Park, Bruce Xiaohan Nie and Song-Chun Zhu
	\IEEEcompsocitemizethanks{
		\IEEEcompsocthanksitem Seyoung Park is with the Department of Computer Science, University of California, Los Angeles. E-mail: seypark@cs.ucla.edu
		\IEEEcompsocthanksitem Bruce Xiaohan Nie is with the Department of Statistics, University of California, Los Angeles. E-mail: xhnie@stat.ucla.edu
		\IEEEcompsocthanksitem Song-Chun Zhu is with the Department of Statistics and Computer Science, University of California, Los Angeles. E-mail: sczhu@stat.ucla.edu.
		}
}

\IEEEtitleabstractindextext{%
\begin{abstract}
	This paper presents an attribute and-or grammar (A-AOG) model for jointly inferring human body pose and human attributes in a parse graph with attributes augmented to nodes in the hierarchical representation. In contrast to other popular methods in the current literature that train separate classifiers for poses and individual attributes, our method explicitly represents the decomposition and articulation of body parts, and account for the correlations between poses and attributes. The A-AOG model is an amalgamation of three traditional grammar formulations: (i) {\em Phrase structure grammar} representing the hierarchical decomposition of the human body from whole to parts; (ii) {\em Dependency grammar} modeling the geometric articulation by a kinematic graph of the body pose; and (iii) {\em Attribute grammar} accounting for the compatibility relations between different parts in the hierarchy so that their appearances follow a consistent style. The parse graph outputs human detection, pose estimation, and attribute prediction simultaneously, which are intuitive and interpretable. We conduct experiments on two tasks on two datasets, and experimental results demonstrate the advantage of joint modeling in comparison with computing poses and attributes independently. Furthermore, our model obtains better performance over existing methods for both pose estimation and attribute prediction tasks.
\end{abstract}

}

\maketitle

\IEEEdisplaynontitleabstractindextext

%
\IEEEpeerreviewmaketitle

\IEEEraisesectionheading{\section{Introduction}\label{sec:introduction}}

%
%
%
%
\subsection{Objectives and Motivations}

\IEEEPARstart{I}{n} this paper, we present a probabilistic, compositional and graphical model for explicitly representing human poses, parts, and attributes in an attribute And-Or grammar (A-AOG), which combines three conventional grammar formulations:
\begin{itemize}
	\item {\em A phrase structure grammar} representing the hierarchical decomposition from whole to parts;
	\item {\em A dependency grammar} modeling the geometric articulation by a kinematic graph of the body pose; and
	\item {\em An Attribute grammar} accounting for the compatibility relations between different parts in the hierarchy so that their appearances follow a consistent style.
\end{itemize}
\begin{figure}[ht!]
	\begin{center}
		\includegraphics[width=250px]{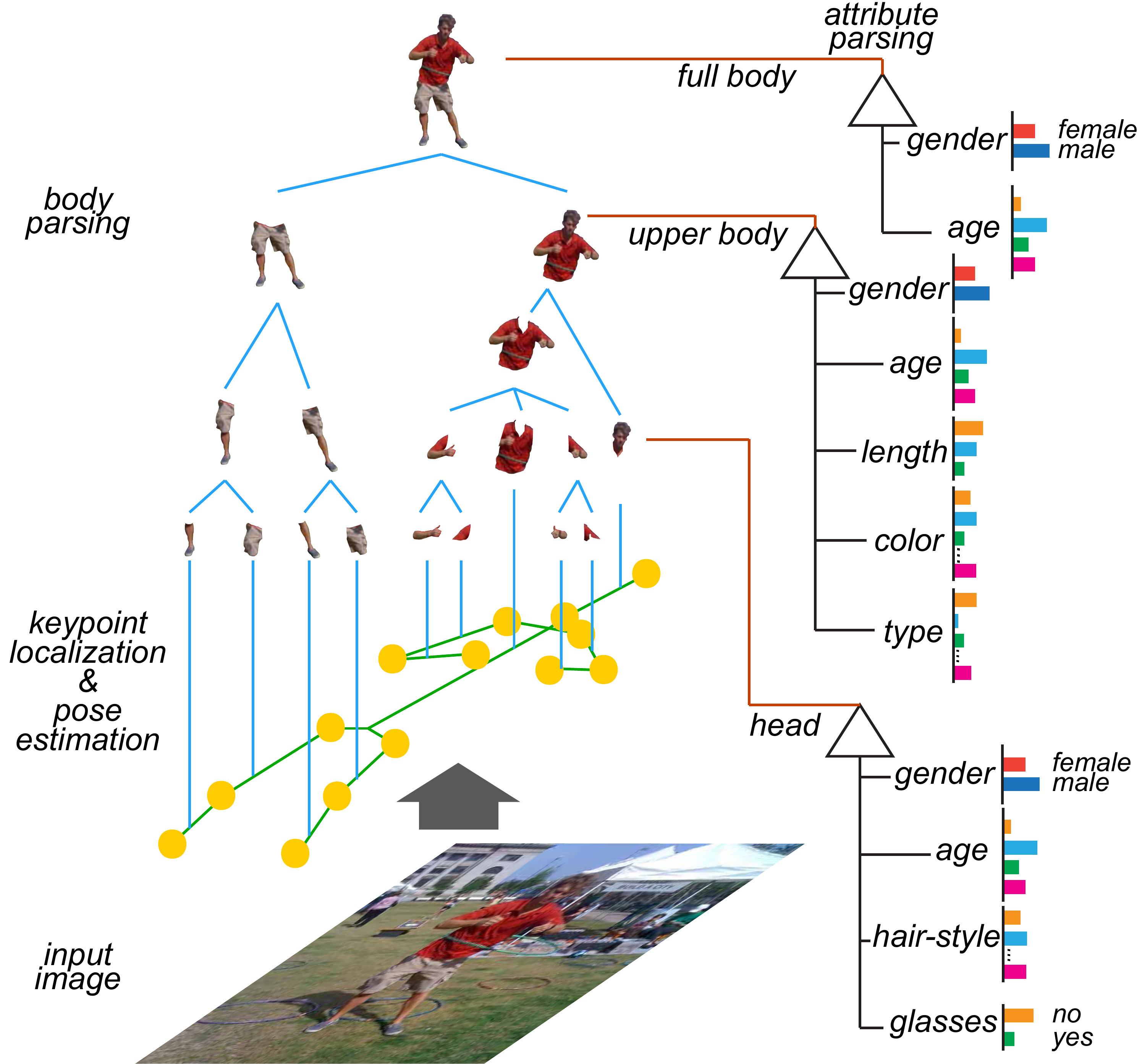}
    	\label{fig:outline}
		\caption{An attributed parse graph for a human image includes three components in colored edges: (i) The hierarchical whole-part decomposition (blue); (ii) the articulation of body parts (green); and (iii) the attributes associated with each node in the hierarchy (red). The parse graph also includes the probabilities and thus uncertainty at each node for attributes.}
	\end{center}
	\vspace{-.5cm}
\end{figure}%

As Figure~\ref{fig:outline} illustrates, our algorithm parses an input image using the A-AoG and outputs an attribute parse graph with three components. (i) The phrase structure parse graph is illustrated by the vertical blue edges, and defines a valid hierarchical composition of body parts following the phrase structure grammar. (ii) The kinematic graph is illustrated by the horizontal green edges, and describes the articulations of body parts following a dependency grammar. (iii) A number of attributes (triangles) are associated with each node in the parse graph as illustrated with the red edges. Each attribute takes values from a finite set, namely its {\em semantic domain}, for example, `gender' $\in$ \{`male', `female'\}, and `hair style' $\in$ \{`long', `short', `bald' \} with posterior probabilities shown in the colored bars.

\begin{figure*}
    \centering
	\includegraphics[width = 510px]{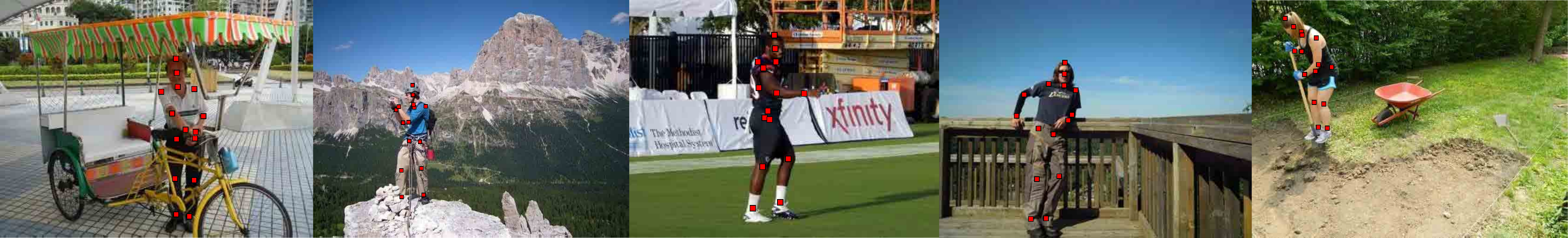}
	\vspace{-.25cm}
	\caption{Examples of our pedestrian attribute dataset. Each image includes one target person and keypoint annotations. This dataset consists of many kinds of variations in pose and attributes. The attribute categories are shown on Table \ref{table:datastat}.}
	\label{fig:ucla_dataset_ex}
	\vspace{-0.5cm}
\end{figure*}

In this representation, some attributes are said to be {\em global}, such as gender and age, as they are associated with all nodes in the parse graph. A global attribute can be inferred from individual body parts, for example, one may tell a person's age or gender from the head, face, or upper body alone with different probabilities, which are illustrated by the colored bars at those nodes. In contrast, the {\em local attributes} are confined to nodes in the low levels of the hierarchy, for example, hair style is an attribute of the head, and can be inferred from the head region alone when the body is occluded. During the inference process, both the global and local attributes pass information top-down and bottom-up in the phrase structure parse graph, and impose constraints to ensure consistency between the parts in a probabilistic way through extra energy terms modeling the correlations between parts. The final output for attribute prediction aggregates information from all the parsed parts.

The following two aspects motivate the proposed A-AOG for jointly modeling and inferring human poses and attributes in an explicit representation.

Firstly, it is desirable to have an explicit and interpretable model for integrating all the attributes in the hierarchical representation. Despite extensive research on attributes for objects~\cite{farhadi2009describing}, humans~\cite{bourdev2011describing}, and scenes~\cite{patterson2012sun} in the past decade, most methods train one classifier for each attribute, such as CNN classifiers, using features extracted from the whole image. As human pose has large variations, it is unreasonable to predict the attributes without knowing where the body parts are, especially for those local attributes. The problem will be more prominent when testing data are significantly different from the training data. Such blackbox classifiers are separated from the well-known graphical models for human pose estimation~\cite{ramanan2006pose, felzenszwalb2010object, rothrock2013integrating} and thus lack explicit understanding of the interactions between attributes and parts. Therefore, an explicit and simple model linking attributes to poses and parts is long desired but missing in the literature.

\begin{figure}[t!]
	\centering
	\includegraphics[width=230px]{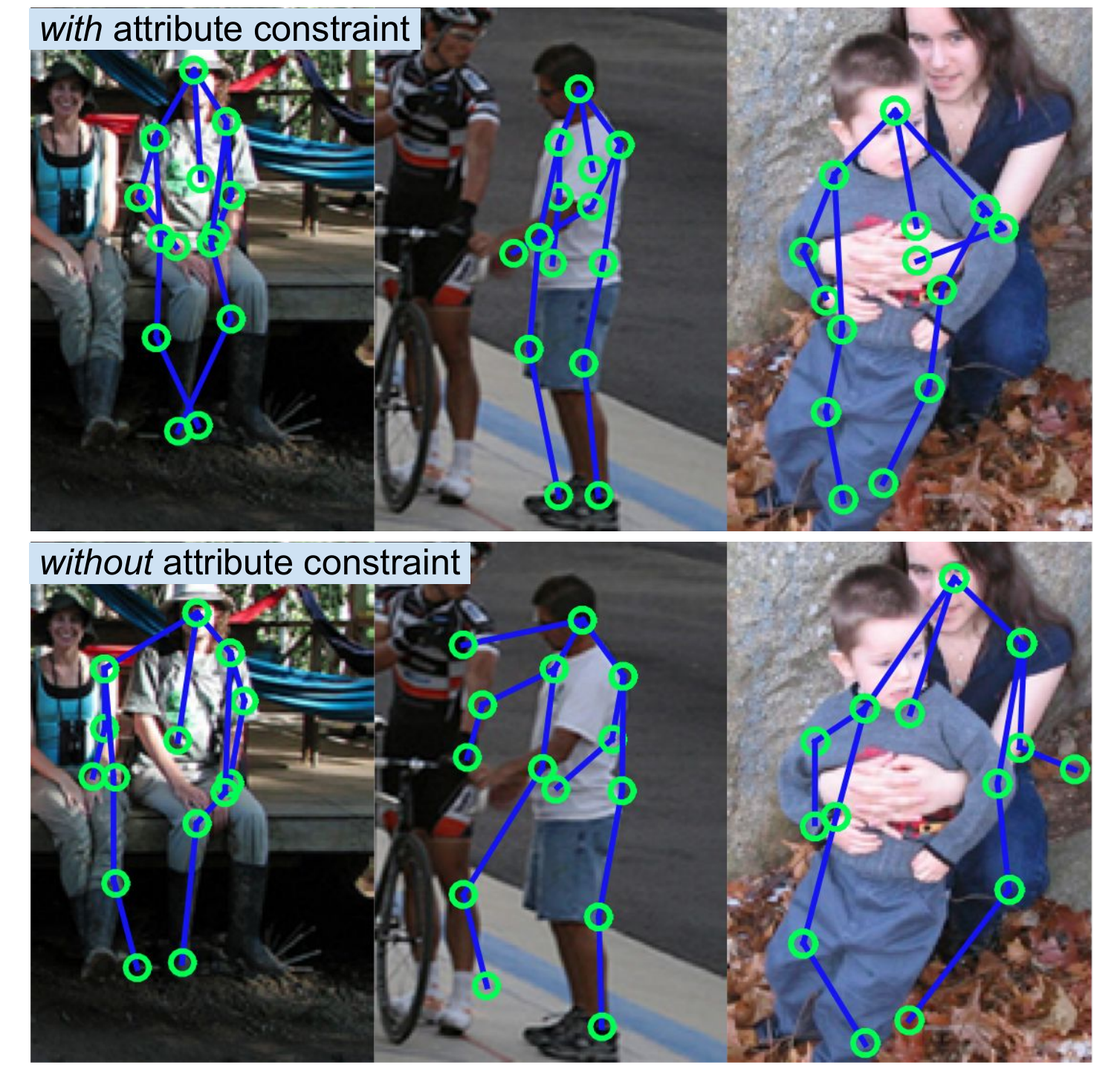}
	\vspace{-.40cm}
	\caption{\textbf{Attribute constraints.} We infer the pose $pt^*$ either using Equation \ref{eqn:pt_wo_attr} or \ref{eqn:pt_w_attr}. When we do not have attribute constraints (i.e. using Equation \ref{eqn:pt_wo_attr}) the model selects the part that maximize the scores in local, and the part could come from different person, when there are multiple people close. However, by having the attribute as global constraints (i.e. using Equation \ref{eqn:pt_w_attr}), we can enforce model to have consistent attributes which subsequently results in better pose-estimation. }
	\label{fig:attr_constraint}
	\vspace{-.45cm}
\end{figure}

Secondly, it is desirable to infer pose and attributes jointly in a common representation. Some recent works use a pre-trained pose detection module~\cite{ramanan2006pose} as a pre-processing stage and then predict attributes sequentially.
Such method inevitably propagates the errors in pose estimation to attribute prediction. We also notice that one significant source of errors in pose estimation comes from nearby person in the image, but the attributes consistency help keep the parts to the same person as seen in Fig.~\ref{fig:attr_constraint}. The joint inference approach solves pose and attributes in an iterative closed-loop and thus utilize the mutual information between pose and attribute, such as the co-occurrence of attributes, and the correlation between attributes and parts.

The proposed A-AOG is aimed at addressing the two aspects above. The A-AOG is a context-sensitive grammar embedded in an And-Or graph~\cite{zhu2006stochastic}. In the and-or graph, the and-nodes represent decomposition or dependency; and the or-nodes represent alternative choices of decomposition or types of parts. The attributes at different levels modulate the choices at the Or-nodes and thus impose the context in a probabilistic way. For example, a female will have higher probability to wear a skirt than a male, and the skirt is represented by some templates in some or-node branch. Near the leaf node, the attribute is directly related to the choices of an Or-node. For example, glasses $\in$ $\{$`yes', `no'$\}$ is a local attribute for head, and a head with or without glasses corresponds to different part templates (or detectors) under an or-node.

\begin{table}[t!]\footnotesize
	\caption{Attribute list in our pedestrian attribute dataset.}\vspace{-.4cm}
	\def\arraystretch{1.4}%
	\def\tabcolsep{10.4 pt}
	\begin{tabular}{l l} \hline
		attributes & semantic domains \\ \hline \hline
		gender & male, female\\
		age	& youth, adult, elderly\\
		hair-style & long-hair, short-hair, bald\\
		upper cloth type & t-shirt, jumper, suit, no-cloth, swimwear\\
		upper cloth length & long-sleeve, short-sleeve, no-sleeve\\
		lower cloth type & long-pants, short-pants, skirt, jeans\\
		glasses& yes, no \\
		hat & yes, no \\
		backpack & yes, no \\ \hline
	\end{tabular}
	\label{table:datastat}
\end{table}

\subsection{Dataset Overview and Scope of Experiments}

We evaluate the performance on multiple benchmarks in various experiment settings.

\textbf{Public dataset on attribute prediction}. The attribute of people dataset~\cite{bourdev2011describing} is a widely contested benchmark. In this dataset, each image is centered and ground truth bounding box of each person is provided with 9 binary attributes. We tested the A-AOG model on attribute classification task in comparison with existing approaches~\cite{joo2013human, zhang2013panda, zhang2013deformable, gkioxari2015actions, gkioxari2015rstarcnn}, and on pose estimation task in comparison with the state-of-the-art method~\cite{chen2014articulated}. We also compared with an approach in conference version~\cite{park2015attributed}. The A-AOG model achieves the state of art performance on both attribute classification and pose estimation tasks.

\textbf{Self-collected dataset for joint parsing}. We collected a pedestrian attribute dataset which was introduced in ICCV 2015~\cite{park2015attributed} for both attribute prediction and pose-estimation tasks. Fig.~\ref{fig:ucla_dataset_ex} shows a
few example images and this dataset has large variation
in margin, size, attribute, pose, appearance, geometry and
environment. We list the attribute categories of this dataset in Table \ref{table:datastat}. We tested the A-AOG model on attribute classification task in comparison with~\cite{gkioxari2015rstarcnn} and a method in our conference version~\cite{park2015attributed}. The results demonstrate that human pose and attributes are highly related and that a joint approach is required for better representation and performance.

\subsection{Contributions}

This paper makes the following contributions:

\begin{itemize}
	\item We propose a novel A-AOG model which combines phrase structure grammar, dependency grammar, and attributed grammar, and is a simple, explicit and interpretable representation in comparison to the neural network models used in~\cite{zhang2013panda,gkioxari2015actions,gkioxari2015rstarcnn}.
	\item We represent pose-estimation, part-localization and attribute prediction problems in a common A-AOG and solve them through joint parsing. This has the following desired properties:
	(i) We represent object appearance, geometry and attributes in a unified model and solve the three tasks simultaneously; (ii) We use a single trained model for multiple attribute predictions, unlike previous approaches that require $n$ models or classifiers for $n$ attributes; (iii) We do not need any pre-processing such as human detection or part localization.
	\item The experiments on widely used Attributes of people dataset~\cite{bourdev2011describing} and Pedestrian attribute dataset~\cite{park2015attributed} show that our method outperforms other state of the art methods on both attribute classification and pose estimation tasks, and demonstrate the benefits and strength of our joint modeling of the two tasks.
\end{itemize}

In comparison to the conference version~\cite{park2015attributed}, this paper provides more details in comparing the three types of grammars and their connections to our A-AOG. We also introduce an efficient way to incorporate deep learned features, and provide extended and improved experimental results with deeper analysis.

The rest of the paper is organized as follows: Section~\ref{sec:relatedworks} summarizes previous works related to our model. Section~\ref{sec:convgrammars} overviews the three types of grammar models. Section~\ref{sec:grammar} explains the proposed attribute and-or grammar modeling. Section~\ref{sec:inference} and~\ref{sec:learning} present the inference and learning algorithms, respectively. Finally, we present various experiments and analysis of experiment results in Section~\ref{sec:experiment}.

\section{Related works  \label{sec:relatedworks}}

Our approach is related to three streams of research in the literature which we will briefly discuss in the following.

\subsection{Research on attribute grammar}
Attribute grammars were first developed in computer science by D. Knuth for formalizing the semantics of context-free languages~\cite{knuth1990attr}, e.g. in compiler writing, and were adopted by natural language processing (NLP)~\cite{abney1997stochastic}. It augments the productions rules in context-free grammars by providing context sensitive conditions on the attributes of the symbols. In computer vision, attribute grammar was adopted and extended by Han and Zhu~\cite{han2009bottom} for parsing man-made scenes. The buildings and furniture, floor, etc. in such scenes are decomposed into rectangles with geometric attributes, such as center, and orientations, which follow some constraints, such as alignment in 1D (windows), 2D (floor tiling), and 3D (cubic objects). In contrast to the hard deterministic constraints in compiler writing, the attribute grammar in scene parsing uses soft constraints in the form of energy terms. Later, the attribute grammar is adopted for action representations~\cite{lin2009semantic, damen2012explaining}, scene attribute tagging~\cite{wang2013weakly}, and 3D scene construction from a single view~\cite{liu2014single}. The grammar rules are manually designed in the scene parsing work~\cite{han2009bottom} and 3D reconstruction work~\cite{liu2014single}. The scene attribute tagging work~\cite{wang2013weakly} learns the attribute grammar in a weakly supervised way from well-aligned images for each scene categories. In contrast, human images have huge pose variations, and thus the pose estimation is integrated into our approach for joint parsing.

\subsection{Research on human attribute classification}
In the past five years, the study of attribute classification has became a popular topic in computer vision for its practical importance. Early work focused on facial images since face is the most informative and distinct part of the body, and is the most suitable for estimating attributes such as gender~\cite{cottrell1990empath, golomb1991sexnet, moghaddam2002learning}, age~\cite{kwon1999age}, and some local attributes (e.g., hair style and glasses)~\cite{kumar2011describable}. Later, as more diverse attributes (e.g., cloth types and actions) were explored, full body parts were used to collect richer and more diverse information. However, input images cannot be used directly without dealing with the variations of geometry and appearance as the full body has large pose variations. Bourdev {et al}. introduced a method to classify attributes in~\cite{bourdev2011describing} by detecting important parts of the body using Poselets~\cite{bourdev2009poselets}, while Chen {et al}. proposed a method to explore human clothing styles with a conditional random field (CRF) in~\cite{chen2012describing} using pre-trained pose estimation\cite{yang2011articulated} output. As these methods used the pre-trained part localization method as a pre-processing step for the sequential steps, the attribute recognition performance undoubtedly relies on the localization accuracy. Joo {et al}. designed a rich appearance part dictionary to capture large variations of geometry and pose~\cite{joo2013human}. Zhang et al. made a considerable performance improvement in~\cite{zhang2013panda} by using the CNN-based approach. They used the pre-trained HOG based poselet approach for part detection and trained classifier with the shallow convolutional neural network for each poselet. This method also relied on part-based approaches and required the ground-truth bounding box at test time. More recently, Gkioxari {et al}. made significant performance improvement in~\cite{gkioxari2015actions, gkioxari2015rstarcnn}. In~\cite{gkioxari2015actions}, three body parts are defined, and jointly fine-tuned with CNN features within the whole body bounding box. In~\cite{gkioxari2015rstarcnn}, the CNN networks are designed to learn context information for attributes and actions. However, this kind of CNN-based approaches still lack the explicit representation.

\subsection{Research on part localization}
Localization and detection of human and its parts has been a topic of interest for many years. The pictorial structure model is introduced in the early stage for detection~\cite{fishler1973therepresentation} and extended in~\cite{andriluka2009pictorial,eichner2009better, felzenszwalb2005pictorial, sapp2010adaptive} which used a geometry-based tree model to represent the body. Since then, the deformable part model~\cite{pedro2010cascade} has became one of the most dominant methods in recent years for detecting humans and other objects~\cite{girshick2011object}. In the last few years, hierarchical mixture models~\cite{rothrock2013integrating, yang2011articulated, yang2013articulated, pishchulin2013strong} made significant progress which is similar to a dependency grammar. Poselets method~\cite{bourdev2009poselets} used a part-based template to interpolate a pose. K-poselets~\cite{gkioxari2014kposelets} improved performance by using poselets as part templates in a DPM model and was based on CNN features~\cite{girshick2014rich}. \cite{chen2014articulated, johnson2011learning, Tompson2014Joint, toshev2014deeppose} showed significant improvement compared to previous methods by training keypoint specific part detectors based on a deep convolutional neural network for human body pose estimation. However, these models did not incorporate any notion of visual attributes explicitly. In comparison, we provide a method for finding informative parts for attributes and their geometrical relations in our model.

\section{Background on Grammar Models	\label{sec:convgrammars}}
In this section, we overview the three types of grammars in the literature to provide the necessary background information and then derive the A-AOG as a unification in the following section.

\subsection{Phrase structure grammar}
The phrase structure grammar, also known as constituency grammar, is based on the constituency relation. The constituency relation defines the rule to break down a node (e.g. parent node) into its constituent parts (e.g. child nodes). In other words, each node must geometrically contain all of its constituents. Phrase structure grammars were introduced in syntactic pattern recognition by K.S. Fu in the early 1980s~\cite{fu1982syntactic}, and rejuvenated into compositional models by Geman et al.~\cite{geman2002composition} and stochastic image grammars by Zhu and Mumford~\cite{zhu2006stochastic}. In the past decade, such grammars have been successfully used in parsing objects and scenes in various form~\cite{zhu2006stochastic, fidler2006hierarchical, rothrock2013integrating}. For example the deformable parts models~\cite{felzenszwalb2010object} is a simple context free phrase structure grammar.

\begin{figure}[t!]
	\begin{center}
		\includegraphics[width=250px]{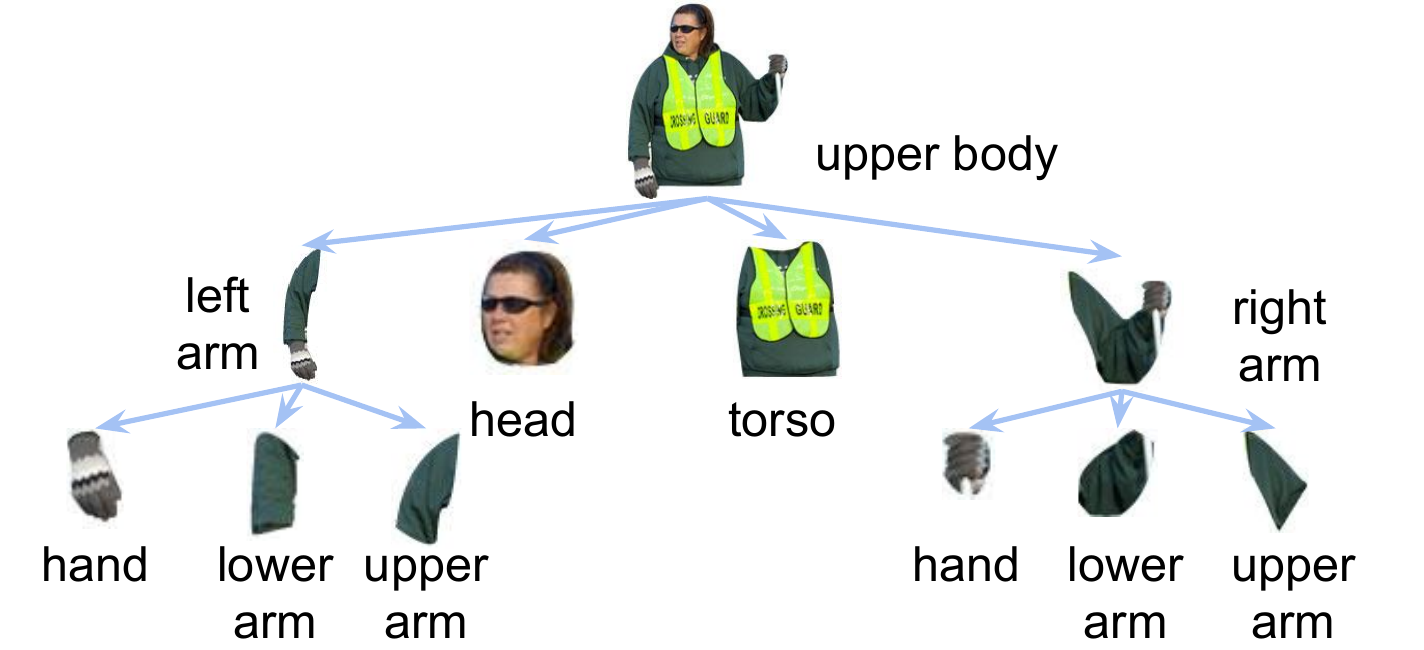}
		\caption{Phrase structure grammar defines coarse-to-fine representation. Each grammar rule decomposes a part into smaller constituent parts.}
		\label{fig:phrase}
	\end{center}
	\vspace{-.4cm}
\end{figure}

Formally, a phrase structure grammar is often formulated as a stochastic context free grammar:
\[ \mathcal{G} =<S, V_N, V_T, R, \mathcal{P}> \]
with five components: a root/initial state $S$, a set of non-terminal nodes $V_N$, a set of terminal node $V_T$, a set of productions rules $R$, and a probability system $P$ associated with these rules. Each terminal node in $V_T$ is represented by an image template or discriminatively trained detector. A typical nonterminal node $A\in V_N$ is derived by some production rules in the following form,
\[
A \rightarrow \alpha_1 \, | \,   \alpha_2 \, | \alpha_3, \quad {\rm with} \; \theta_1 | \theta_2 | \theta_3.
\]
where $\alpha_i$ is a string of nodes in $V_N \cup V_T$ and $\theta_i$ is the branching probability for the three distinct ways for deriving node $A$. Fig.~\ref{fig:phrase} illustrates a parse graph example. The root node is the upper body and decomposed into arms, head, and torso. The arms are further decomposed into upper arm, lower arm, and hand. It can be described by production rules:
\begin{eqnarray*}
	\textrm{torso} 			& \rightarrow & \textrm{l. arm, head, torso, r. arm} \\
	\textrm{l. arm}		 	& \rightarrow & \textrm{l. upper arm, l. lower arm, l. hand} \\
	\textrm{r. arm}			& \rightarrow & \textrm{r. upper arm, r. lower arm, r. hand}
\end{eqnarray*}

l. and r. indicate left and right respectively. In a general form, the grammar can derive a large number of parse graphs for human images depending on the clothing styles, body poses and camera views. Such grammar models have at least two shortcomings in vision tasks.
\begin{itemize}
	\item It lacks contextual information, such as the correlations between sibling parts or the conditions for expanding an non-terminal node.
	\item It is often disadvantageous to choose only one of the branches exclusively due to ambiguities in image appearance or because the image templates under different branches are not well-separable. Thus it loses performances to implicit and noisy models like the convolutional neural net.
\end{itemize}

\subsection{Dependency grammar}
Dependency grammars have been widely used in natural language processing for syntactic parsing, especially for languages with free word order. It has a root node $S$ and a set of $n$ other nodes $\{A_1, ..., A_n\}$ with production rules like
\begin{eqnarray}
S & \rightarrow & A_1 \, | \, A_2 \cdots | A_n;\\
A_i &\rightarrow& a_i \, | \, a_i A_j \, | \, A_j a_i;  \quad \forall i=1,2...,n, j\neq i.
\end{eqnarray}

The root node can transit to any other node once, and then each node $A_i$ can terminate as $a_i$ or transit to another node $A_j$ to the left or right side. Unlike phrase structure grammars, a child node derived from a parent node is not a constituent of the parent but depends on the parent in some semantic relations.

\begin{figure}[t!]
	\begin{center}
		\includegraphics[width=250px]{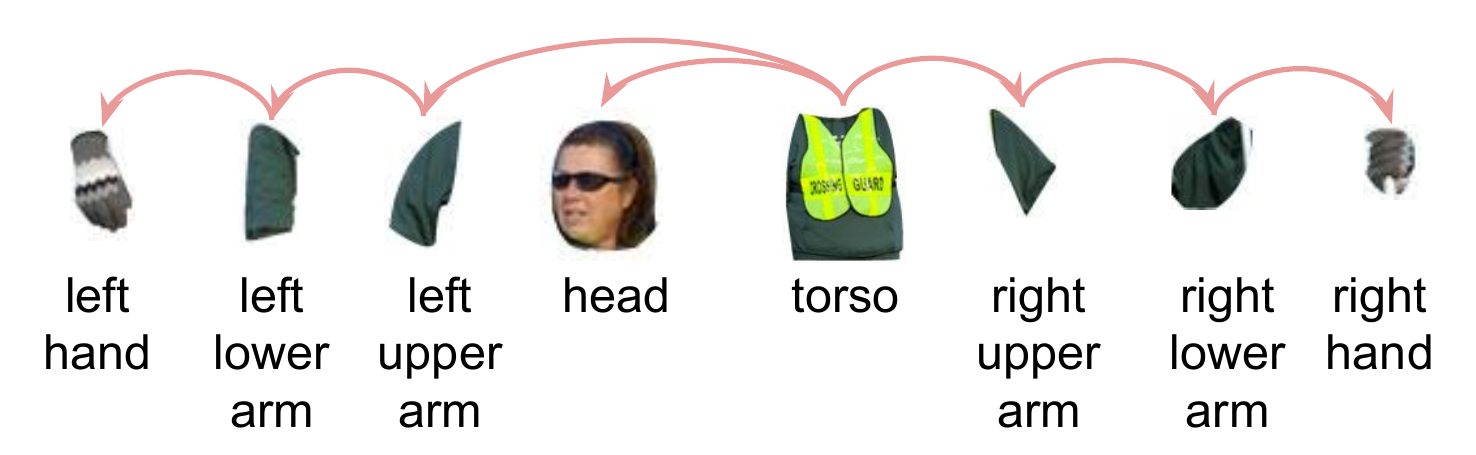}
		\caption{Dependency grammar defines adjacency relations that connects the geometry of a part to its dependent parts. It is well suited for representing objects that exhibit large articulated deformations. }
		\label{fig:dependency}
	\end{center}
\end{figure}

For example, Fig.~\ref{fig:dependency} is a parse graph for the upper body derived from a dependency grammar and is called a kinematic parse graph. The root node is the torso part as it is the center of the body and connected to other parts. The upper arms and head are the child nodes of the torso. It can be described with production rules as:
\begin{eqnarray*}
	\textrm{torso} 			& \rightarrow & \textrm{l. upper arm, head, r. upper arm} \\
	\textrm{l. upper arm} 	& \rightarrow & \textrm{l. lower arm} \\
	\textrm{l. lower arm}	& \rightarrow & \textrm{l. hand} \\
	\textrm{r. upper arm}	& \rightarrow & \textrm{r. lower arm} \\
	\textrm{r. lower arm}	& \rightarrow & \textrm{r. hand}
\end{eqnarray*}

The dependency grammar is well suited for representing objects that exhibit large articulated deformations.
Body parts at different locations and orientations are treated as different nodes in this grammar. In computer vision, the pictorial model~\cite{felzenszwalb2005pictorial} and the flexible mixture of parts model~\cite{yang2013articulated} can be viewed as dependency grammars.
The advantage of the dependency grammar lies in its simplicity which facilitates learning. The disadvantage is that it loses the coarse-to-fine summarization of the phrase structure grammar.

\subsection{Attribute grammar}

Attribute grammar assigns some attributes $\{x_1, x_2, ...,x_k\}$ to the non-terminal or terminal nodes of a grammar, e.g. A non-terminal node $A$ (or terminal node $a$) has attributes denoted by $A.x_1$ (or $a.x_1$).	These attributes have semantic domains as we shown in Table~\ref{table:datastat}. Such assignments macth our intuition in visual perception.

Then for each production rule in a context-free grammar, we can augment a number of conditions or constraints between the attributes of parents and children nodes in the following form,
\begin{eqnarray}
A & \rightarrow & a B c  \\
&s.t.&  A.x_1 = f_1(a.x_1, B.x_2, c.x_1) \nonumber \\
& & A.x_2 = f_2(a.x_2, B.x_1, c.x_2). \nonumber
\end{eqnarray}
In the above rule, the functions $f_1$ and $f_2$ impose constraints and conditions for deriving the node $A$, so that the grammar becomes mildly context-sensitive. Furthermore, they provide means for passing information top-down (for the so-called inherited attributes) and bottom-up (for the so-called synthesized attributes) in the parse graph.

Our A-AOG relaxes the hard constraints \cite{han2009bottom,liu2014single} to soft energy terms and encodes three types of contextual information.
\begin{itemize}
	\item Consistency between the same attribute in parent and children nodes, for example, if the root node has gender attribute as female, then its parts are likely also female.
	\item Co-occurrence between attributes, e.g. a female is more likely to have long hair and wear a skirt.
	\item Correlations between the assignment of an attribute to the image feature of a node and its alternative choices, e.g. long hair and short hair will have preferences on the choices of image templates.
\end{itemize}

\section{Attribute And-Or Grammar Model \label{sec:grammar}}

In this section, we present the A-AOG to integrate the three types of grammars into an And-Or graph representation~\cite{zhu2006stochastic}.

\subsection{Attribute And-Or Graph Representation}
We construct the And-Or graph in three steps:
\begin{enumerate}
	\item We use a phrase structure grammar as the backbone of the And-Or graph, which is compositional and reconfigurable, i.e. its parse graph can change structures in contrast to some hierarchical models with fixed structure.

	\item We augment the hierarchy with dependency relations, i.e. edges, to encode the articulations between adjacent body parts.

	\item We further associate the nodes in the And-Or graph with attributes, which expand the dimensions (or variables) of representation, and introduce additional contextual constraints.
\end{enumerate}

\begin{figure*}[ht!]
	\begin{center}
		\includegraphics[width=520px]{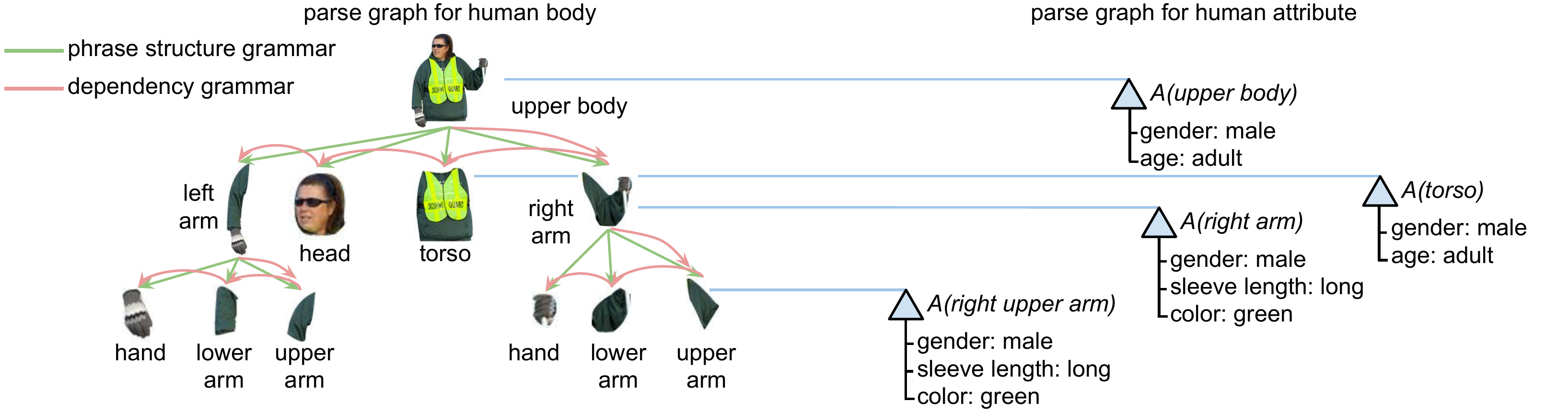}
		\vspace{-.45cm}
		\caption{An example of the parse graph following the A-AOG, which includes a parse graph for human body detection and pose and a parse graph for human attributes.}
		\label{fig:aaog_parse}
	\end{center}
	\vspace{-.6cm}
\end{figure*}

For clarity, we denote the A-AOG by a 5-tuple:
\[ A-AOG =<S, V, E, X, \mathcal{P}>. \]

1) The vertex set $V=V_{\rm and} \cup V_{\rm or} \cup V_T$ consists of three subsets: (i) a set of and-nodes $V_{\rm and}$ for decomposition in the phrase structure grammar; (ii) a set of or-nodes $V_{\rm or}$ for branching to alternative decompositions and thus enabling reconfiguration; and (iii) a set of terminal nodes $V_T$ for grounding on image features. Each node $v\in V$ represents a body part of different granularities and has state variables designating the location $(x,y)$. In fact, the state variables are the geometric attributes which are propagated between parent-children nodes in the hierarchy. However, in this paper, we treat them separately from the human attributes.

\begin{figure*}[t!]
	\begin{center}
		\includegraphics[width=510px]{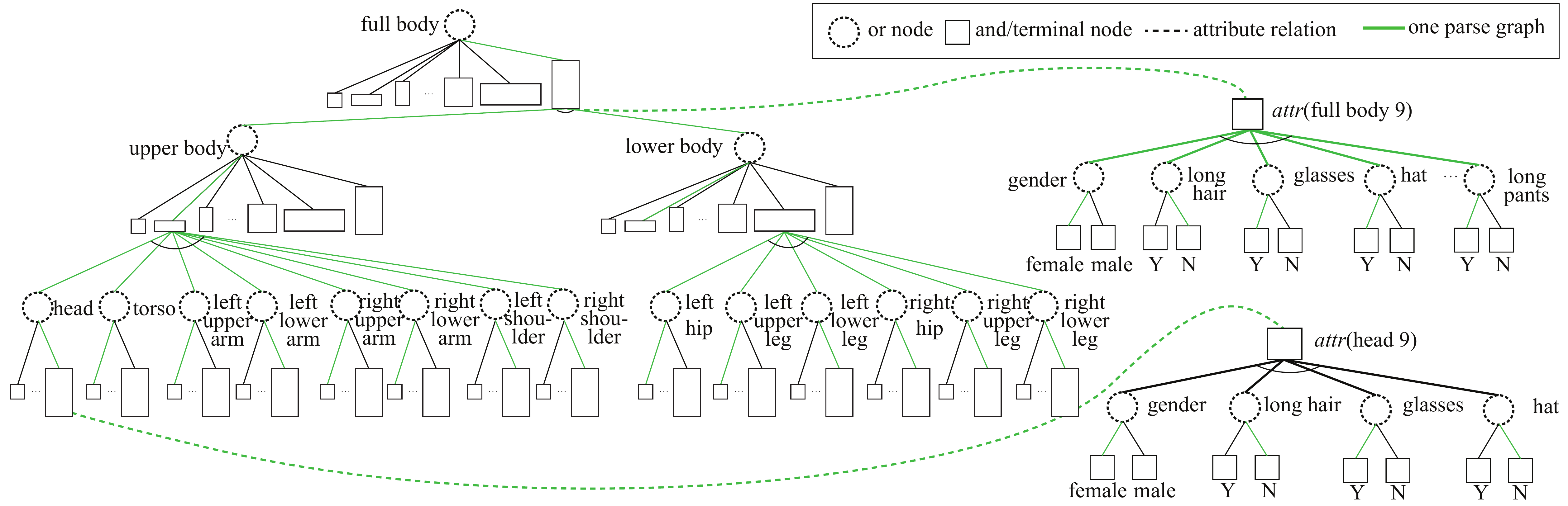}
		\vspace{-.5cm}
		\caption{We define 14 atomic parts, 2 mid-level parts, and root part. We define 9 part types for each part. Each attribute has corresponding parts, and connected with part types through attribute relations. In this figure, we only show attributes of head and full body, and draw phrase structure grammar relation for better illustration.}
		\label{fig:aaog}
	\end{center}
	\vspace{-.6cm}
\end{figure*}

We define 14 atomic, i.e. terminal, parts, head, torso, l.shoulder, r.shoulder. l.upper arm, l.lower arm, r.upper arm, r.lower arm, l.hip, r.hip, l.upper leg, l.lower leg, r.upper leg, and r.lower leg. These parts are defined as terminal nodes ($V_T$) in the grammar. We then define non-terminal nodes ($V_T$), upper body and lower body by combining terminal parts. Upper body part includes head, torso, shoulders, and arms. Lower body part includes hips and legs. The root part, full body, is defined by upper body and lower body. We illustrate the defined grammar in Fig.~\ref{fig:aaog}.

2) The edge set $E=E_{\rm psg} \cup E_{\rm dg} $ consists of two subsets: (i) a set of edges with phrase structure grammar $E_{\rm psg}$; and (ii) a set of edges with dependency grammar $E_{\rm dg}$.

3) The attribute set $X=\{x_1, ...,x_9\}$ are associated with nodes in $V$.

4) $\mathcal{P}$ is the probability model on the graphical representation.

Define the parse graph
\[  pg = (V(pg), E(pg), X(pg))
\]

where $V(pg)$, $E(pg)$ and $X(pg)$ are the set of nodes, edges and attributes respectively in parse graph $pg$. Fig.~\ref{fig:aaog_parse} shows a parse graph example derived from the A-AOG, which includes a parse graph for human body detection and pose and a parse graph for human attributes.

\subsection{Formulation of Joint Pose and Attribute Parsing}
The goal is to find the most probable parse graph $pg$ from the designed grammar model given an image $I$. The probability model over the parse graph is formulated in a Bayesian framework, which computes the joint posterior as the product of a likelihood and prior probability, and equivalently represented as the following the Gibbs distribution
\begin{align}
P(pg|I;\lambda) &\propto P(I|pg;\lambda)P(pg;\lambda) \nonumber\\
&= \frac{1}{Z} \exp\{-\mathcal{E}(I|pg;\lambda) -\mathcal{E}(pg;\lambda)\} \label{eqn:energy}
\end{align}

The model parameters are denoted as $\lambda$. The energy functions $\mathcal{E}$ are further decomposed into a set of potential functions. These potentials constrain all aspects of the grammar. The likelihood term describes appearance response and is decomposed into part and attribute appearance.

\begin{align}
-\mathcal{E}(I|pg;\lambda) = -\mathcal{E}_{\textrm{app}}^V(I|pg;\lambda) -\mathcal{E}_{\textrm{app}}^{X}(I|pg;\lambda)
\end{align}

\noindent
$\mathcal{E}_{\textrm{app}}^{V}(I|pg;\lambda)$ and $\mathcal{E}_{\textrm{app}}^{X}(I|pg;\lambda)$ are appearance terms for part and attribute respectively. The prior term is used to describe relations in grammar. It is also decomposed into part and attribute relations.

\begin{align}
-\mathcal{E}(pt;\lambda) = -\mathcal{E}_{\textrm{rel}}^{V}(pg;\lambda)-\mathcal{E}_{\textrm{rel}}^{X}(pg;\lambda)\}
\end{align}

\noindent
$\mathcal{E}_{\textrm{rel}}^{V}(pg;\lambda)$ and $\mathcal{E}_{\textrm{rel}}^{X}(pg;\lambda)$ are relation terms for part and attribute respectively. We rewrite Equation \ref{eqn:energy} as
\begin{align}
P(pg|I;\lambda) = {\frac{1}{Z}} \exp \{&-\mathcal{E}_{\textrm{app}}^V(I|pg;\lambda)-\mathcal{E}_{\textrm{app}}^{X}(I|pg;\lambda) \nonumber\\
&-\mathcal{E}_{\textrm{rel}}^{V}(pg;\lambda)-\mathcal{E}_{\textrm{rel}}^{X}(pg;\lambda)\}
\end{align}

We, then, can express energy terms as scoring functions.
\begin{align}
S(pg,I) &= -\mathcal{E}_{\textrm{app}}^{V}(I|pg)-\mathcal{E}_{\textrm{app}}^{X}(I|pg)-\mathcal{E}_{\textrm{rel}}^{V}(pg)-\mathcal{E}_{\textrm{rel}}^{X}(pg) \nonumber \\
&= S_{\textrm{app}}^{V}(I,pg) + S_{\textrm{app}}^{X}(I,pg) + S_{\textrm{rel}}^{V}(pg) + S_{\textrm{rel}}^{X}(pg)\nonumber \\
\end{align}

We now have four scoring functions.

\begin{eqnarray*}
&S_{\textrm{app}}^{V}(I,pg) & \textrm{part appearance score function}\\
&S_{\textrm{app}}^{X}(I,pg) & \textrm{attribute appearance score function }\\
&S_{\textrm{rel}}^{V}(pg)   & \textrm{part relation score function}\\
&S_{\textrm{rel}}^{X}(pg)   & \textrm{attribute relation score function}
\end{eqnarray*}

The choice and particular forms for these scoring functions vary on the design and intention of the grammar, which we explore in the following sections.

\subsection{Part Model}

\subsubsection{Part Relation Model}
The term $S_{\textrm{rel}}^{V} (pg)$ is for the part relation score. We define two relation types: syntactic relation (green edges in Fig.~\ref{fig:aaog_parse}) and kinematic relation (red edges in Fig.~\ref{fig:aaog_parse}). Syntactic relation follows phrase structure grammar, and controls the part composition. Kinematic relation follows dependency grammar rule, and describes the articulation constraints. 
The overall relation score sums each of these relations.


Each part $v_i$ is associated with $(x_i,y_i)$ as its position and $t_i$ as its part type. The syntactic relation is defined between parts to represent the co-occurrence frequency of two neighboring part types, and captures correlations and compatibilities between the parts. It is described by score function $S_{\textrm{rel}}^{S}(v_i, v_j)$, and the $S_{\textrm{rel}}^{S}$ over parse graph is computed as:
\begin{align}
S_{\textrm{rel}}^S(pg) 	= 	\sum_{(i,j) \in E_{\textrm{psg}} (pg)} S_{\textrm{rel}}^S(v_i, v_j)
\end{align}
\noindent
where $E_{\textrm{psg}}(pg)$ is the set of edges with phrase structure grammar in parse graph $pg$. Here $S_\textrm{rel}^S(v_i,v_j)=logP(t_i,t_j)$ where $P(t_i,t_j)$ is the probability that part type $t_i$ occurs with $t_j$.

%
%

The kinematic relation is defined between parts by and-rule from and-or grammar. And-rule can be viewed as the rule for assembly of constituent parts and enforce geometric constraints between two parts ($v_i$, $v_j$) with relative geometry of articulation. It is described by score function $S_{\textrm{rel}}^{K}(v_i, v_j)$.
%
%

$S_{\textrm{rel}}^K$ over the parse graph is computed as
\begin{align}
S_{\textrm{rel}}^K(pg) 	= 	\sum_{(i,j) \in E_{\textrm{dg}} (pg)} S_{\textrm{rel}}^K(v_i, v_j)
\end{align}

\noindent
where $E_{\textrm{dg}}(pg)$ is the set of edges with dependency grammar in parse graph $pg$. $S_\textrm{rel}^K(v_i,v_j)=logP(v_i,v_j)$ where $P(v_i,v_j)$ is the probability of the mixture of Gaussian on $(x_i-x_j,y_i-y_j)$. We write part relation score function $S_{\textrm{rel}}^V(pg)$ as below.\\
\begin{align}
S_{\textrm{rel}}^V(pg) 	&= 	S_{\textrm{rel}}^S(pg) + S_{\textrm{rel}}^K(pg) \nonumber \\
&=	\sum_{(i,j) \in E_{\textrm{psg}} (pg)} S_{\textrm{rel}}^{S}(v_i, v_j)  + \sum_{(i,j) \in E_{\textrm{dg}} (pg)} S_{\textrm{rel}}^K(v_i, v_j)
\end{align}

\subsubsection{Part Appearance Model}
In the previous section, we defined parts and its relations. We now design appearance templates to describe diverse appearances under different viewpoint and pose for part and its types. Appearance is described by the image likelihood of the grammar and corresponding to the scoring function $S_{\textrm{app}}^V(pg, I)$, and it can be computed from the part appearance score function $S_{\textrm{app}}^{V}(v_i,I)$ which indicates the local appearance evidence for placing $i$-th part $v_i$ on image patch centered at location $(x_i,y_i)$. $S_{\textrm{app}}^{V}$ over $pg$ is computed as:
\begin{align}
S_{\textrm{app}}^V(pg,I) 	&= 	\sum_{v_i \in V(pg)} S_{\textrm{app}}^V(v_i, I)
\end{align}

\noindent
$V(pg)$ is a set of parts in parse graph $pg$.

\subsection{Attribute Model \label{sec:attributemodel}}
We now combine attribute notation on the grammar model by defining relations between part and attributes. Previous attribute approaches~\cite{bourdev2011describing, chen2012describing, joo2013human, zhang2013deformable, zhang2013panda} use all defined parts (or poselets) for attribute classifications, and it means they assumed attributes are related to all body parts. However, we can simply know some parts may not be related with such attributes, and it might hurt attribute prediction if we classify attribute with unrelated parts. For examples, `glasses' is not related to `lower body' parts and `t-shirt' is not related to `head; or `lower body' parts. In contrast, `long-hair' attribute will be highly related to `head' part. Therefore, we need to learn how attributes and parts are related. We define the set of attributes for each part $v$ and denote them by $X(v)$. Then, $X(v)$ includes related attributes for part $v$. As we illustrated in Fig.~\ref{fig:aaog_parse} and Fig.~\ref{fig:aaog}, we attach $X(v)$ to each part in our grammar. We will discuss how to obtain $X(v)$ in Section~\ref{sec:mi_learning}.

\subsubsection{Attribute Node}
We can treat the set of attributes $X(v)$ for part $v$ as a two layered simple graph which follows and-or grammar rules as illustrated on the right side in the Fig.~\ref{fig:aaog}. The root node of $X(v)$ is described by and-rule in and-or grammar. It includes corresponding attributes for part $v$ as its child nodes. Then, each attribute includes attribute types as child nodes, it follows or-rule. It has two attribute types for binary attribute class, such as `gender' and `wearing t-shirt', or have more than two for multi-class attributes, such as `cloth types' or `age'. It can be described in production rule. For example, when part $v$ has two corresponding attributes, `gender' and `wearing t-shirt', `gender' can have types `male' and `female'. `wearing t-shirt' has two child nodes, `yes' and `no'. In production rule, it could be written as follow,
\begin{table}[h!]
    \centering
	\footnotesize
	\begin{tabular}{c c}
		attribute production rule & example \\
		{$\!\begin{aligned}
			X(v) 	&\rightarrow \{X_1, X_2\}\\
			X_1 	&\rightarrow X_{11} | X_{12} \\
			X_2 	&\rightarrow X_{21} | X_{22} \\
			\end{aligned}$}
		&
		{$\!\begin{aligned}
			X(v)	&\rightarrow \{\textrm{Gender}, \textrm{T-shirt}\}\\
			\textrm{Gender} 	&\rightarrow \textrm{Female} | \textrm{Male} \\
			\textrm{T-shirt}		&\rightarrow \textrm{Yes} | \textrm{No} \\
			\end{aligned}$}
	\end{tabular}
    \vspace{-.5cm}
\end{table}
%
%
%
%
\subsubsection{Attribute Relation Model \label{sec:attr_relation}}
Each attribute node $X(v_i)$ is linked to part $v_i$ through a relation as shown by blue edges in Fig.~\ref{fig:aaog_parse} and a dashed line in Fig.~\ref{fig:aaog}. This relation describes properties of part node $v_i$ and reflects the co-occurrence frequency of the attribute given the part type. For example, let the specific part type of node $v$ (= upper body) have an appearance that is blouse-like. This will occur more frequently with female than male, and therefore the model should favor selecting this part when there is strong evidence for the female attribute. It is described by score function $S_{\textrm{rel}}^{X} (v_i, X(v_i))$, and $S_{\textrm{rel}}^{X}$ over $pg$ is computed as
\begin{align}
S_{\textrm{rel}}^X(pg) = 	\sum_{v_i \in V(pg)} S_{\textrm{rel}}^X(v_i, X(v_i))
\end{align}

\subsubsection{Attribute Appearance Model}

Just as we defined appearance model for part, so too we define appearance model for attribute. It corresponds to the scoring function $S_{\textrm{app}}^{X}(pg, I)$, and it can be computed from attribute score function $S_{\textrm{app}}^{X}(X(v_i), I)$. It indicates the local appearance response of attributes of part $v_i$ at image path centered at $v_i = (x_i, y_i)$. The score $S_{\textrm{app}}^X$ over $pg$ is computed as
\begin{align}
S_{\textrm{app}}^X(pg, I) & = \sum_{v_i \in V(pg)} S_{\textrm{app}}^X(X(v_i), I)
\end{align}
\subsection{Combine Appearance Models}

We defined two appearance models: part appearance $S_{\textrm{app}}^V(v_i, I)$ and attribute appearance $S_{\textrm{app}}^X(X(v_i), I)$. They are now connected through the part-attribute relation, and we can combine those two appearance score functions into single function $S_{\textrm{app}}(v_i, a_j, I)$ where $a_j \in X(v_i)$. In order to capture the diverse appearance of part and attributes under different viewpoints and poses, we borrow the strength from a deep CNN model. At the last layer of our CNN model, we can directly get $P(v_i, a_j, t_m| I)$ which is the likelihood of the image patch that belongs to part $v_i$ with part type $t_m$ and attribute $a_j$. The score function $S_{\textrm{app}}(v_i, a_j, I) = \log(P(v_i, a_j, t_m, I))$. The total appearance score over $pg$ is computed as:
\begin{align}
S_{\textrm{app}}(pg, I) = \sum_{v_i\in V(pg)} \sum_{a_j \in X(v_i)} S_{\textrm{app}}(v_i, a_j, I)
\end{align}

We will explain how we utilize a CNN model to obtain part and attribute scores together in Section \ref{sec:learning}.

\section{Parsing and Inference \label{sec:inference}}

We defined grammar structure for describing human body with attributes. The inference task is now equivalent to finding the most probable parse graph $pg^*$ from the constructed grammar model for given image $I$. We can find $pg^*$ by maximizing the score functions described in previous sections.
\begin{align}
pg^{*} 	&= \arg \max_{pg} P(I|pg) P(pg) \nonumber \\
&= \arg \max_{pg} [S_{\textrm{app}}^V(pg, I) + S_{\textrm{app}}^X(pg, I) + S_{\textrm{rel}}^V(pg) + S_{\textrm{rel}}^{X}(pg)]
\label{eqn:total_score_separate}
\end{align}

We combine part appearance score $S_{\textrm{app}}^V(pg, I)$ and attribute score $S_{\textrm{app}}^X(pg, I)$ into $S_{\textrm{app}}(pg, I)$. And, the part and attribute relation $S_\textrm{rel}^X(pg)$ will also be captured by $S_{\textrm{app}}(pg,I)$ because we model them in the same CNN model. Equation \ref{eqn:total_score_separate} can be rewritten as:
\begin{align}
pg^{*} 	&= \arg \max_{pg} [S_{\textrm{app}}(pg, I) + S_{\textrm{rel}}(pg)]
\label{eqn:total_score_cnn}
\end{align}

We now denote part relation score function $S_{\textrm{rel}}^{V}(\cdot)$ by $S_{\textrm{rel}}(\cdot)$ to simplify the equation. In the conference version~\cite{park2015attributed}, we used Equation~\ref{eqn:total_score_separate} to find the optimal parse graph $pg^*$, but we maximize the Equation~\ref{eqn:total_score_cnn} in this paper. We compare two methods quantitatively in Section~\ref{sec:experiment}.

In the first step of inference, in order to reduce the searching space of the parse graph, we generate the proposals for each part by the deep CNN model instead of computing response maps on image pyramid. Each part proposal is associated with attributes.
After having the proposals for each part: $O_{v_1}, O_{v_2},...,O_{v_n}$, we can obtain the final parse graph $pg^*$ with Equation \ref{eqn:total_score_cnn}. To maximize $S(pg, I) = S_{\textrm{app}}(pg, I) + S_{\textrm{rel}}(pg)$, we introduce the total score function $S_{\textrm{tot}}(v_i, I)$ which is formulated in a recursive way:
\begin{align}
S_{\textrm{tot}}(v_i,I) 	&= \sum_{j=1}^A S_{\textrm{app}}(v_i, a_j, I) \nonumber \\
&\quad+ \sum_{v_j \in C(v_i)} [S_{\textrm{rel}}(v_i, v_j) + S_{\textrm{tot}}(v_j, I)]
\label{eqn:pt_wo_attr}
\end{align}
where A is the number of attribute categories and $C(v_i)$ is the set of children of part $v_i$, and $v_i \in O_{v_i}$. The score function $S(pg, I)$ is equivalent to $S_{\textrm{tot}}(v_o, I)$ where $v_o$ is the root part. We now can infer the parse graph $pg^*$ by maximizing $S_{\textrm{tot}}(v_o, I)$, and it can be expressed as
\begin{align}
pg^* = \arg \max_{pg} S_{\textrm{tot}}(v_0, I) \label{eqn:argmax_no_attr}
\end{align}
\noindent

We can also infer the parse graph $pg^*$ in a different way by using the attribute as a global constraint which means all parts should have the same attribute. We first find attribute specific parse graph $pg_{a_j}$. $pg_{a_j}$ denotes the parse graph under $j$-th attribute $a_j$. To infer parse graph $pg_{a_j}^*$, we maximize $S(pg_{a_j}, I) = S_{\textrm{app}}(pg_{a_j}, I) + S_{\textrm{rel}}(pg)$ where $S_{\textrm{app}}(pg_{a_j}, I) = \sum_{v \in V(pg_{a_j})} S_{\textrm{app}} (v, a_j, I)$ . To maximize $S(pg_{a_j}, I)$, we use score function $S_{\textrm{tot}}(v_i, a_j, I)$
\begin{align}
S_{\textrm{tot}}(v_i, a_j, I) 	&= S_{\textrm{app}}(v_i, a_j, I) \nonumber \\
&\quad+ \sum_{v_j \in C(v_i)} [S_{\textrm{rel}}(v_i, v_j) + S_{\textrm{tot}}(v_j, a_j, I)]
\label{eqn:pt_w_attr}
\end{align}

\noindent
The score function $S(pg_{a_j}, I)$ is now equivalent to $S_{\textrm{tot}}(v_0, a_j, I)$, and the inference of parse graph can be expressed as $pg_{a_j}^* = \arg \max_{pg_{a_j}} S_{\textrm{tot}}(v_0, a_j, I)$. Commonly this maximization problem can be computed using dynamic programming (DP), however, the DP cannot be used because of many loopy cliques in our model due to the combination of phrase structure grammar and dependency grammar. Here we applied a greedy algorithm based on the beam search.

We start with the parse graph which only includes one root part $v_0$ and generate $N_o = |O_{v_0}|$ parse graph candidates $pg_1',...,pg_N'$ from the proposals $O_{v_0}$. The score of each parse graph candidate $S(pg_i')$ is computed as $S_{app}(v_0,a_j,I)$, then one child part of the root part $v_i \in O_{v_i}$ is added into parse graph and generate $N_1 = N_0 \times |O_{v_i}|$ parse graph candidates. The score $S(pg_i')$ is updated by adding the part relation score $S_{rel}(v_i,v_j)$. Only top-K high scored parse graph candidates are kept into next step. We continue adding child part of current parse graph, updating the scores and pruning candidates until all parts are added into the parse graph. Finally we pick the parse graph candidate with highest score as the inferred parse graph $pg_a^*$.

We have $pg_a^*$ as many as the number of attributes. For example, we have $9 \times 2$ attribute specific parse graphs $pg_{a_j}^*$ where $j = \{1,2,\dots, 18\}$ in the experiment for Attribute of People dataset which define 9 binary attributes. The final parse graph $pg^*$ on image $I$ is inferred as:
\begin{align}
pg^* = \arg \max_{pg_{a_j}} S(pg_{a_j}^*, I)
\end{align}

\begin{figure}[t!]
	\centering
	\includegraphics[width=250px]{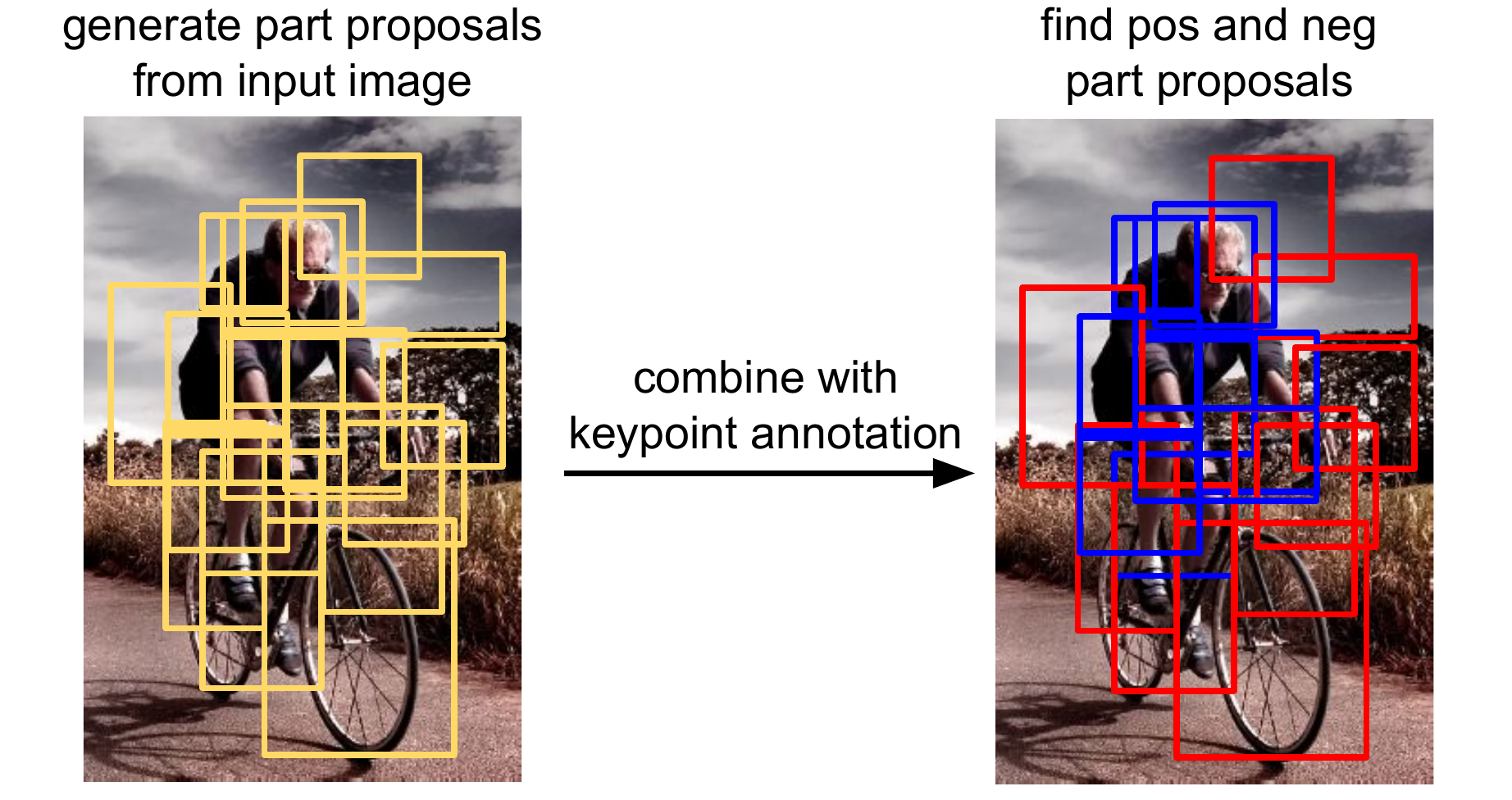}
	\vspace{-.7cm}
	\caption{We generate part proposals from input image and find positive and negative proposals using keypoint annotation information.}
	\label{fig:part_proposal}
	\vspace{-.35cm}
\end{figure}

We compare two pose inference methods in Fig.~\ref{fig:attr_constraint}. As we can see from the figure, when we do not include attribute constraints by using Equation \ref{eqn:pt_wo_attr}, the model selects parts that maximize the scores locally and the parts could come from different people. However, when we have global attribute constraints with Equation \ref{eqn:pt_w_attr}, we can control the model to have consistent attributes and can provide better pose estimation.

We compute the score of attribute $S(a_j)$ from each parse graph $pg_{a_j}^*$ and it is the summation of part score on the parse graph $pg_{a_j}^*$ under part-attribute constraints.
\begin{align}
S(a_j) = \sum_{v \in V(pg_{a_j}^*)} S_{\textrm{app}}(v_i, a_j, I) \cdot \mathbbm{1}(a_j, X(v_i)) \label{eqn:attr_score}
\end{align}

\noindent
$X(v_j)$ is the set of attributes corresponding to part $v_i$ and defined in Section \ref{sec:attr_relation}. $\mathbbm{1}(\cdot)$ is an indicator function.

\section{Learning \label{sec:learning}}
Our learning procedure contains two stages: the first stage learns deep CNN models for part proposals generation, and computation of the likelihood of part and attribute, and part positions regression; the second stage learns the geometric relations between parts, and the compatibility relations between parts and attributes.

\subsection{Part proposal and appearance model learning}
\label{sec:part_proposal}

Although Faster R-CNN~\cite{ren2015fasterrcnn} is used for generic object detection task, it has not been used in the task of fine-grained object recognition and detection, e.g., attribute classification and part localization.~\cite{ren2015fasterrcnn} trained two networks: region proposal network for object proposal generation and fast-rcnn network~\cite{girshick2015fastrcnn} for object recognition. Inspired by the success in~\cite{ren2015fasterrcnn}, we design two networks: part proposal network to generate part proposals; the part-attribute network to generate likelihood for all part-attribute combinations using classification layer and predict positions of parts using a regression layer. The two networks share weights of bottom layers. The two networks are initialized with the same pre-trained models in~\cite{ren2015fasterrcnn}.

\begin{figure}[t!]
	\centering
	\includegraphics[width=250px]{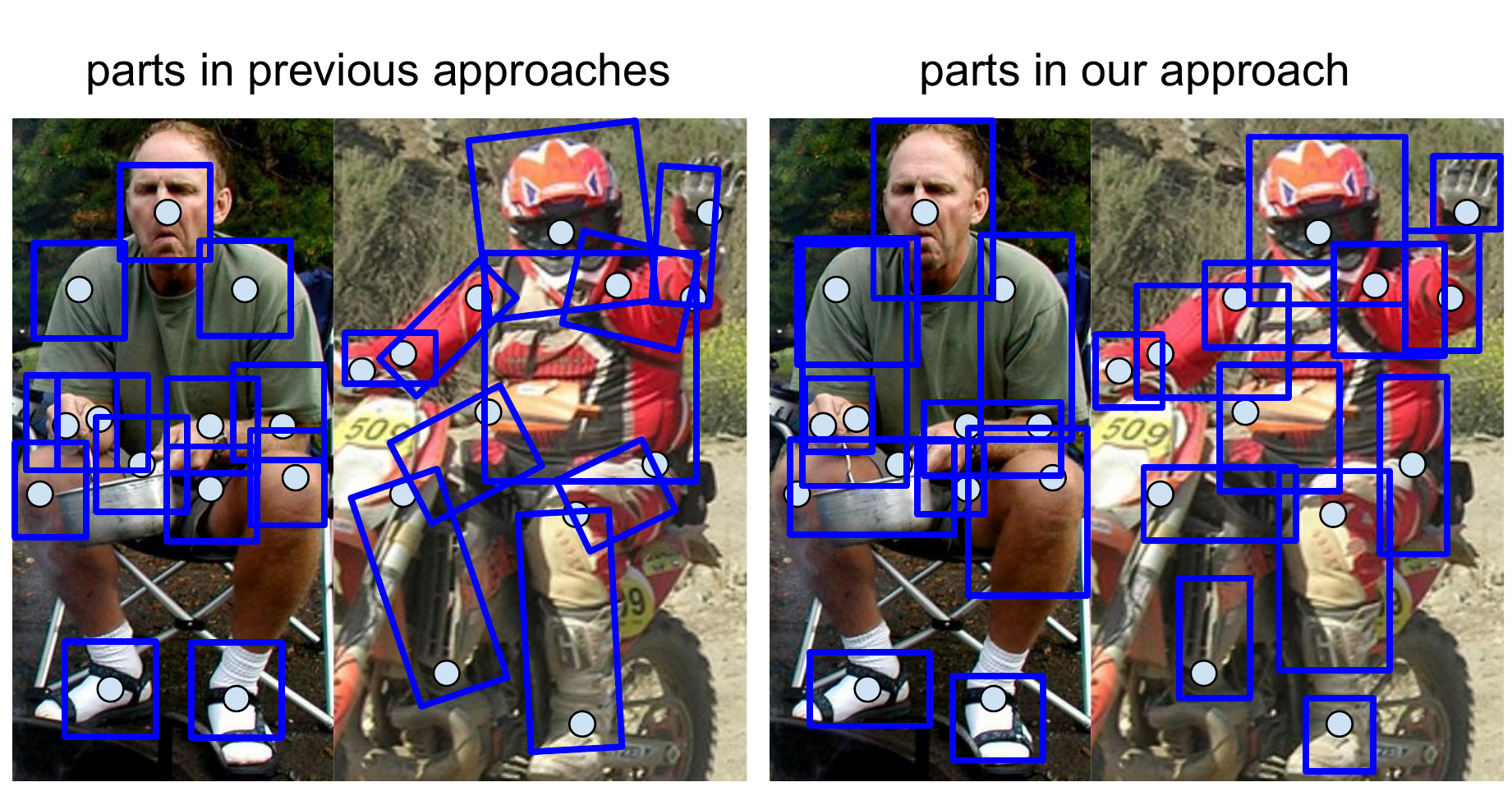}
	\vspace{-.85cm}
	\caption{In previous approaches, (\textit{left}) parts are defined by drawing square bounding box around keypoint~\cite{yang2013articulated, chen2014articulated} or by annotating precise bounding box~\cite{rothrock2013integrating}. (\textit{right}) We define parts based on our part proposal process, and it handles geometric and scale variation of parts efficiently and effectively.}
	\label{fig:part_define}
	\vspace{-.35cm}
\end{figure}

We use strong supervision which includes the 14 joints of human (see Figure~\ref{fig:result_poselet_pose}), the bounding box of the target person and the attribute labels. To train the part proposal network, we first generate 9 bounding boxes ($=$ 3 scales $\times$ 3 aspect ratios) which correspond to 9 part types at each location, and then compute the overlap between 9 bounding boxes and the ground-truth bounding box for each location. The locations are predicting the part proposals if the overlap is bigger than 0.5. From the part proposal network, we can generate proposal set $O_i$ for each part $v_i$. The part type $t_p \in \{1,2,...,9\}$ for each proposal is the index of the predicted bounding box  with highest probability. To train part-attribute network, we decide the training labels for each proposal by the following process: (1) compute the overlap between the proposed and the ground-truth human bounding box; (2) label the proposals with overlap lower than 0.5 for negatives. (3) select the proposals with overlap higher than 0.7 and compute the minimum distance $d_k$ between each proposal and all parts as $\min_{i=1}^{n}(||[x_k,y_k]-[x_i,y_i]||^2 / \min(w_k,h_k))$. $[x_k,y_k]$ is the center of a proposal $k$, $[x_i,y_i]$ is the keypoint of part $i$, and $[w_k,h_k]$ are the width and height of the proposal, and $n$ is the total number of parts. The keypoint of atomic part is defined as its joint and the keypoint of a non-terminal part (upper-body, lower-body, full-body) is the center of joints included in this part. We also record the part index $I_p$ which gives the minimum distance. $I_p$ is the index of a non-terminal part if the proposal includes all the joints of this part, or the index of an atomic part otherwise. (4) keeps the proposals whose $d_k$ are smaller than $0.5$. Each proposal is labeled as the part index $I_p$ and part type $t_p$. We illustrate the part proposal process in Fig.~\ref{fig:part_proposal} and compare our part design with previous approaches in Fig.~\ref{fig:part_define}. From our approach for part design, we can handle large variation of part scale and aspect ratio.

In part-attribute network, the output of the classification layer is the likelihood of all part-attribute combinations for each proposal region. The length of the output is $17 \times 9 \times k + 1$ in which 17 is the number of parts (14 atomic parts + 2 mid-level parts + root part) and 9 is the number of part types and $k$ is the number of attribute categories, and +1 is for negatives (i.e. background). The output of regression layer are the keypoint positions of parts.

\subsection{Part relation and part-attribute relation learning \label{sec:mi_learning}}

The part relation includes the syntactic relation between parent and children parts, and kinematic relation between atomic parts. The syntactic relation is defined as $S_{rel}^S(v_i,v_j)=logP(t_i,t_j)$ and $P(t_i,t_j)$ is from the normalized histogram of co-occurrence of part types. We use the mixture of Gaussian to measure the part relation $P(v_i,v_j)$ and the score $S_{\textrm{rel}}^K(v_i,v_j)=logP(v_i,v_j)$, and it penalizes the displacement of the $i$-th part and $j$-th part. The number of mixtures is set to 10.

\begin{figure}[t!]
	\centering
	\includegraphics[width=250px]{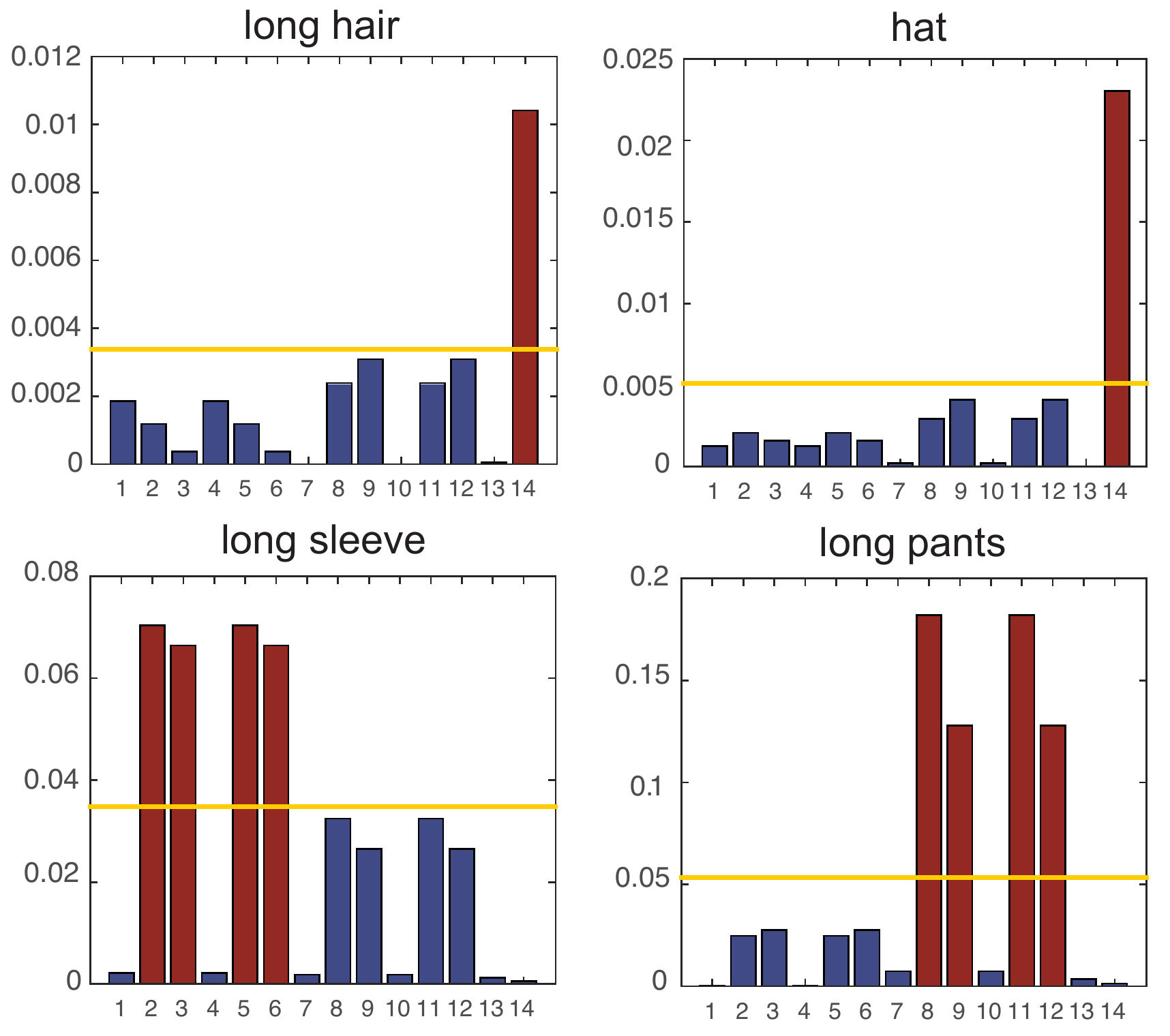}
	\vspace{-.7cm}
	\caption{\textbf{Part-attribute relation}. We compute mutual information between atomic parts and attribute to find the parts that contribute the most to each attribute. If the mutual information value is higher than the mean value (the yellow lines), we consider the part to be associated with the attribute (the red bars). We only compute for 14 terminal parts. The mid-level and root parts will synthesize the attribute relations from child nodes. Please see text for details and part indexes. }
	\label{fig:mi}
	\vspace{-.35cm}
\end{figure}

The part-attribute relation describes the compatibility relationship between part and attribute. In the Attributes of People dataset~\cite{bourdev2011describing}, about one third of the images are annotated unknown for each attribute, and many joints are also annotated invisible due to the occlusion or truncation of human body. In most cases, the attributes are annotated unknown because of the invisible specific parts. For example, the attribute '{long pants}' of most images are unknown because only the upper body is visible. Thus, these annotations actually provide strong evidence about which parts play a role in distinguishing the attribute. To find associated part for each attribute, we compute mutual information between attributes with label `{known}' and `{unknown}' and 14 atomic parts with label `{visible}' and `{invisible}'. We show the computed mutual information in Fig.~\ref{fig:mi} for four attributes. For each attribute $a_j$, we pick the parts of which the values are above the mean value as the associated part set. If a part is picked then its parent part will also be picked. For example, `{leg}' is the related part for `{jeans}', and `{lower body}' and `{full body}' are considered as related parts for `{jeans}' as well. We denote corresponding attributes for each part by $X(v)$.

\section{Experiments \label{sec:experiment}}

We conduct three sets of experiments on joint inference of human attribute and pose. The first set of experiments evaluates our method on attribute classification, and compares against the previous approaches in~\cite{joo2013human, zhang2013panda, park2015attributed, gkioxari2015rstarcnn, gkioxari2015actions}. The second set tests pose estimation, and compares against the state-of-the-art method in~~\cite{chen2014articulated}. The last set is diagnostic analysis and compares our joint model with its different variants and settings. In particular, we show that pose estimation improves attribute classification and vice versa.
\\

\subsection{Benchmarks}
\noindent
\textbf{Attributes of People dataset.} Introduced by~\cite{bourdev2011describing}, this datataset consists of 4013 training images and 4022 testing images. Each image is centered at the target person which is annotated as a bounding box. This dataset defines 9 binary attributes, and keypoint annotations can be used for training. This dataset is challenging for attribute classification and pose estimation because the person is always occluded and truncated, and the pose variation is very large.\\

\noindent
{\textbf{Pedestrian Attribute dataset.} We collected and annotated a Pedestrian attribute dataset. It provides part and attribute annotations as illustrated in Figure \ref{fig:ucla_dataset_ex}. It is designed to have one
person in an image, and does not provide a bounding box of the target human at test time. It includes 2257 high resolution images of which 1257 images are for training and 1000 images for testing. It consists of many types of variations in attribute, pose, appearance, geometry and environment. The 16 joint positions and labels of 9 attributes are provided.

\begin{table*}
	\scriptsize
	\centering
	\caption{Attribute prediction performance on attributes of people dataset. 8 layers indicates using 8 layer structure CNN model, and 16 layers indicates 16 layer model.}
	\vspace{-.25cm}
	\def\arraystretch{1.5}%
	\def\tabcolsep{9.1pt}
	\begin{tabular}{l c c c c c c c c c c}
		Method & Male & \makecell{Long\\hair} & Glasses & Hat & T-shirt & \makecell{Long\\sleeve} & Shorts & Jeans & \makecell{Long\\pants} & mAP\\ \hline \hline
		\textbf{1. with ground truth bounding box} & & & & & & & & & & \\
		Joo et al.~\cite{joo2013human} & {88.0} & 80.1 & 56.0 & 75.4 & {53.5} & {75.2} & 47.6 & 69.3 & 91.1 & 70.7 \\
		PANDA~\cite{zhang2013panda} & 91.7 & 82.7 & {70.0} & 74.2 & 68.8 & {86.0} & {79.1} & 81.0 & {96.4} & {78.98}\\
		Park et al.~\cite{park2015attributed} & {92.1} & {85.2} & 69.4 & {76.2} & {69.1} & 84.4 & 68.2 & {82.4} & 94.9 & {80.20} \\
		Gkioxari et al. (8 layers)~\cite{gkioxari2015actions} & 91.7 & \textbf{86.3} & 72.5 & 89.9 & \textbf{69.0} & 90.1 & 88.5 & \textbf{88.3} & \textbf{98.1} & 86.0 \\
		Ours w pose (8 layers) & {91.9} & 85.0 & {79.7} & {90.4} & 65.5 & {92.1} & {89.9} & 87.3 & 97.9 & {86.7} \\
		Ours w pose + w or-nodes (8 layers) & \textbf{93.0} & 86.2 & \textbf{80.2} & \textbf{91.8} & 67.1 & \textbf{93.6} & \textbf{91.4} & 88.2 & \textbf{98.1} & \textbf{87.7} \\ \cline{1-1}
		Gkioxari et al. (16 layers)~\cite{gkioxari2015actions} & {92.9} & {90.1} & {77.7} & {93.6} & {72.6} & {93.2} & {93.9} & {92.1} & \textbf{98.8} & {89.5} \\
		R* CNN  (16 layers)~\cite{gkioxari2015rstarcnn} & 92.8 & 88.9 & {82.4} & 92.2 & \textbf{74.8} & 91.2 & 92.9 & 89.4 & 97.9 & 89.2 \\
		Ours w pose (16 layers) & {94.9} & {90.6} & {85.2} & {93.7} & 71.3 & {95.1} & {94.2} & {93.1} & \textbf{98.8} & {90.7}\\
		Ours w pose + w or-nodes (16 layers) & \textbf{95.2} & \textbf{92.0} & \textbf{86.3} & \textbf{94.8} & 72.9 & \textbf{95.9} & \textbf{94.8} & \textbf{93.5} & \textbf{98.8} & \textbf{91.6}\\
		\hline
		\textbf{2. without ground truth bounding box} & & & & & & & & & & \\
		Gkioxari et al. (8 layers)~\cite{gkioxari2015actions} & 84.1 & 77.9 & 62.7 & 84.5 & \textbf{66.8} & 84.7 & 80.7 & 79.2 & 91.9 & 79.2 \\
		Ours w/o pose (8 layers) & {88.3} & {84.1} & 73.2 & 86.4 & 57.1 & 90.1 & 78.8 & 85.1 & 95.8 & 81.6 \\
		Ours w pose(8 layers) & {87.9} & {83.6} & {75.4} & {87.3} & 62.2 & {92.1} & {84.1} & {87.6} & {97.6} & {84.2}\\
		Ours w pose + w or-nodes (8 layers) & \textbf{89.2} & \textbf{85.2} & \textbf{76.3} & \textbf{88.7} & \textbf{63.9} & \textbf{93.5} & \textbf{85.5} & \textbf{89.8} & \textbf{98.1} & \textbf{85.6}\\ \cline{1-1}
		Gkioxari et al. (16 layers)~\cite{gkioxari2015actions} & 90.1 & 85.2 & 70.2 & 89.8 & 63.2 & 89.7 & 83.4 & 84.8 & 96.3 & 83.6 \\
		Ours w/o pose (16 layers) & 92.1 & 88.4 & 76.4 & 90.1 & 62.7 & 92.8 & 82.5 & 89.2 & 98.1 & 85.8\\
		Ours w pose (16 layers) & {93.7} & {91.1} & {78.5} & {92.6} & {68.2} & {94.0} & {88.4} & {92.1} & {98.6} & {88.5}\\
		Ours w pose + w or-nodes (16 layers) & \textbf{94.8} & \textbf{91.9} & \textbf{79.4} & \textbf{93.8} & \textbf{69.1} & \textbf{95.1} & \textbf{89.1} & \textbf{93.2} & \textbf{98.8} & \textbf{89.4}\\
		\hline
		\hline
	\end{tabular}
	\vspace{-.5cm}
	\label{table:poselet_attr}
\end{table*}

\begin{table*}[th!]
	\centering
	\scriptsize
	\caption{Results for attribute classification on the proposed \textbf{Pedestrian attribute dataset}. We use average accuracy for evaluation. The ground truth bounding boxes are not provided at test time. We use the R-CNN~\cite{girshick2014rich} person detection method for input for R* CNN because it requires the person bounding box at test time. We only evaluate binary attributes for R-CNN. Please see text for details.}
	\vspace{-.25cm}
	\bgroup
	\def\arraystretch{1.4}%
	\def\tabcolsep{7.5 pt}
	\begin{tabular}{l  c c c c c c c c c c}
		Method & Gender & Age & Hair-style & \makecell{Upper cloth\\type} & \makecell{Upper cloth\\length} & \makecell{Lower cloth\\type} & Backpack & Glasses & Hat & mAC \\ \hline \hline
		Park et al.~\cite{park2015attributed} & {79.8} & {88.2} & {71.7} & {74.9} & {77.2} & {69.9} & {70.8} & {61.1} & {78.1}   & 74.6 \\
		R* CNN~\cite{gkioxari2015rstarcnn} & 79.5 & - & - & - & - & - & \textbf{90.4} & 84.1 & 84.3 & -\\
		Ours w/o pose & 84.1 & 87.1 & 80.8 & 74.1 & 78.8 & 72.8 & 88.2 & 80.1 & 81.1 & 80.8\\
		Ours w pose & {85.2} & {89.6} & {84.9} & {79.9} & {81.2} & {77.5} & 89.6 & {85.3} & {85.5} & {84.3}\\
		Ours w pose + w or-nodes & \textbf{86.8} & \textbf{90.5} & \textbf{86.0} & \textbf{81.2} & \textbf{82.5} & \textbf{78.7} & 90.2 & \textbf{86.9} & \textbf{87.0} & \textbf{85.5}\\ \hline
	\end{tabular}
	\label{table:ucla_attr}
	\egroup
	\vspace{-.5cm}
\end{table*}

\subsection{Evaluation on Attribute Classification}

Table \ref{table:poselet_attr} compares the average precision of attribute classification between our method and other methods~\cite{joo2013human, zhang2013panda, park2015attributed, gkioxari2015rstarcnn, gkioxari2015actions} on Attributes of People dataset. All the methods from the 1st row to the 10th row use the ground-truth bounding box of the target person at test time. To compute the log likelihood of part and attribute in our structure, we train our 8-layer networks and 16-layer networks based on the same pre-trained networks used in~\cite{ren2015fasterrcnn}. For attribute classification evaluation, we compute the score using Equation \ref{eqn:attr_score}. Note that~\cite{joo2013human} did not use joint annotations during training, and~\cite{zhang2013panda} trained model on a different training dataset (25K images) which is much larger than the original set. Our method with pose (i.e. joint model) outperforms the state-of-the-art methods~\cite{gkioxari2015actions} using 8-layer and 16-layer network architectures respectively.

We also compare our approach when ground truth bounding boxes of the target persons are not given at test time. The results are shown from the 11th row in Table \ref{table:poselet_attr}. In~\cite{gkioxari2015actions}, they use R-CNN~\cite{girshick2014rich} person detection as the preprocessing step to detect the target person. However, our approach detects people, classifies attributes, and estimates human pose simultaneously in a unified framework. Overall, our method with 8-layer network improves mAP of same network architecture by 6.4\% point, and even achieves 2.0\% point better mAP than 16-layer network in~\cite{gkioxari2015actions}. In addition, we achieve 5.8\% point better performance than 16-layer network architecture in~\cite{gkioxari2015actions} using the same network. We also show a tremendous improvement from our conference version~\cite{park2015attributed}. It is important to note that we perform better with the same network architecture, and it demonstrates the advantage of the joint modeling in which pose-estimation and attribute prediction help each other during training and inference.

We compare the attribute classification performance on self-collected Pedestrian attribute dataset on Table \ref{table:ucla_attr}. We outperform our conference version~\cite{park2015attributed} substantially on all attributes without the ground truth boxes at test time. To have a better comparison, we train the model of R* CNN~\cite{gkioxari2015rstarcnn} on this dataset. This dataset is designed without providing ground truth bounding box at test time, but R* CNN requires the ground truth bounding box of target human. In order to have a fair comparison, we use the person detection method in~\cite{ren2015fasterrcnn} to generate bounding boxes for R* CNN instead of using ground truth bounding boxes. We only train and evaluate binary attributes, because R* CNN is designed for binary attribute classification. We show better performance on 3 binary attributes out of 4. We show some examples of our results in Figure~\ref{fig:result_ucla}.

\begin{table*}[th!]
    \scriptsize
	\centering
	\caption{Pose estimation result on the Attributes of People dataset. All the methods above the double horizontal line use the ground-truth bounding boxes of target persons at test time.}
	\bgroup
	\def\arraystretch{1.25}%
	\def\tabcolsep{7.2pt}
	\begin{tabular}{l c c c c c c c c c c c c c c }
		Method & S1 & S2 & S3 & S4 & S5 & S6 & S7 & S8 & S9 & S10 & S11 & S12 & S13 & avg\\ \hline \hline
		\textbf{1. with ground truth bounding box} & & & & & & & & & & & & & &\\
		Chen et al.~\cite{chen2014articulated} & 29.7 & 33.2 & 25.0 & 30.1 & 34.5 & {25.5} & 37.1 & 36.1 & 38.1 & 38.4 & 38.2 & 36.5 & 46.0 & 34.5\\
		Ours w attribute & {71.0} & {49.6} & {26.2} & {71.2} & {50.1} & 25.1 & {57.3} & {48.4} & {44.2} & {57.7} & {49.8} & {45.7} & {80.5} & {52.0} \\
		Ours w attribute + w or-nodes & \textbf{71.9} & \textbf{50.7} & \textbf{28.1} & \textbf{72.8} & \textbf{51.8} & \textbf{26.2} & \textbf{58.4} & \textbf{49.9} & \textbf{45.5} & \textbf{59.0} & \textbf{51.1} & \textbf{47.1} & \textbf{81.4} & \textbf{52.4} \\
		\hline
		\textbf{2. without ground truth bounding box} & & & & & & & & & & & & & & \\
		Chen et al.~\cite{chen2014articulated} & 28.5 & 34.0 & 25.5 & 29.7 & 34.7 & 26.3 & 33.9 & 34.8 & 35.0 & 33.9 & 34.5 & 34.7 & 43.7 & 33.0 \\
		Ours w/o attributes & 69.5 & 45.7 & 23.8 & 69.3 & 45.6 & 25.4 & 48.1 & 39.2 & 35.3 & 49.9 & 40.4 & 38.5 & 76.3 & 46.7\\
		Ours w attributes & {73.6} & {52.6} & {29.2} & {74.3} & {53.2} & {28.5} & {54.2} & {42.7} & {41.5} & {52.5} & {43.9} & {41.5} & {79.4} & {51.3} \\
		Ours w attributes + w or-nodes & \textbf{74.7} & \textbf{54.1} & \textbf{30.4} & \textbf{75.5} & \textbf{54.1} & \textbf{29.7} & \textbf{55.4} & \textbf{44.1} & \textbf{42.4} & \textbf{54.1} & \textbf{45.1} & \textbf{42.9} & \textbf{80.1} & \textbf{52.5} \\
		\hline
	\end{tabular}
	\label{table:poselet_pose}
	\egroup
	\vspace{-.5cm}
\end{table*}

\subsection{Evaluation on Pose Estimation}

We show the pose estimation result on the Attributes of People dataset in Table~\ref{table:poselet_pose}. A widely used evaluation method, strict Percentage of Correct Part (PCP) is used as the evaluation metric to compare with the state-of-the-art method in~\cite{chen2014articulated}. Unlike traditional pose estimation methods which are designed to test on the image with one person and small margin, our approach does not require a bounding box because we detect human and estimate pose. For a better and fair comparison we conduct experiments under two different settings. In the first setting, ground truth bounding boxes are provided at test time, while the second setting does not have such. Overall, our method shows a substantial improvement on both settings. We show part indexes in Fig.~\ref{fig:result_poselet_pose} (b). We believe it is because previous approaches conduct experiments on images with similar scales for human detection, but images in this dataset has large scale and appearance variations, and heavy occlusion and truncation. For instance, the traditional pose estimation dataset in~\cite{ramanan2006pose} is scaled to contain people of roughly 150
pixels in height; however, the smallest height of humans present in this dataset is 72 pixels and the largest is 1118 pixels in this attribute dataset. 

We show examples of pose-estimation in Fig.~\ref{fig:result_poselet_pose} (a). The top row shows successful examples, and the bottom row shows failures. As we can see from the examples, our method handles large variation of human pose, occlusion and truncation very well. Although we improve the pose-estimation in multiple people in the scene using the attribute constraints as shown in Fig.~\ref{fig:attr_constraint}, most of our failures come from the situation with multiple people because it is hard to distinguish part identities when people have the same attributes.

\subsection{Diagnostic experiments \label{sec:analysis}}
To better justify the contribution of joint modeling of attribute and pose, we have two diagnostic comparisons: 1) attribute prediction without pose; 2) pose estimation without attributes.\\

\noindent
\textbf{Attribute prediction without pose.} Instead of inferring the best pose for each attribute and computing the corresponding attribute score, we pick the highest score among all part proposals for each attribute and use that score as the attribute classification score. The mean average precision shown in Table~\ref{table:poselet_attr} is 2.6\% (8-layer) and 2.7\% (16-layer) point lower than the result from our joint model without or-nodes when a ground truth bounding box is not provided. It demonstrates that the attributes can be improved when the pose is inferred together. Impressively, it is still better than~\cite{gkioxari2015actions} with the same 8-layers network architecture because we detect parts with a large number of different scales and aspect ratios instead of 200 poselets used by~\cite{gkioxari2015actions}. We also see similar result when a ground truth bounding box is provided.\\

\noindent
\textbf{Pose estimation without attribute.} We infer the pose using the part proposals from all attributes instead of one pose for each attribute. In other words, we use Equation~\ref{eqn:pt_wo_attr} instead of Equation~\ref{eqn:pt_w_attr}. In this approach, $pt^*$ provides the pose by allowing parts to have inconsistent attributes because we maximize the attribute score for each part. This approach can cause problems at pose estimation by choosing parts from different people who are close to each other. This kind of mistake happens frequently in existing pose estimation approaches, and that's why they require the bounding box of target person at test. In Table~\ref{table:poselet_pose}, the performance of pose estimation is decreased by 4.6\% point from our joint modeling with attribute and without or-nodes. From Fig.~\ref{fig:attr_constraint}, we can clearly see that the parts of the estimated pose are from different person in the image if no attribute information is used. However, this problem is solved to some extent when we infer the best pose for each attribute because different people from the same image may have different attributes.

\section{Conclusion}
This paper presents an attributed and-or grammar to describe compositionality, reconfigurability, articulation, and attributes of human in a hierarchical joint representation. Our approach parses an input image by inferring the human, body pose, parts, and attributes. The advantage of our approach is the ability to perform simultaneous attribute reasoning and part detection, unlike previous attribute models that use large numbers of region-based attribute classifiers without explicitly localizing parts.

Our method currently requires large keypoint annotations, unlike recent approaches did not. The keypoint is required to parse pose and attribute jointly. If we use the small number of parts, e.g. head, upper body, lower body, we can avoid large keypoint annotations but we still can learn relations between part and attribute.

We demonstrate our model on benchmarks, and achieve the state-of-the-art attribute classification and pose estimation performance against recent methods. We also show our joint modeling for part and attribute model has better results on the pose-estimation task than previous pose-estimation approaches. We believe that evidence of the attributes as well as their consistency constraints can lead to performance improvements on both tasks.

\begin{figure*}[th!]
	\begin{center}
		\includegraphics[width=450px]{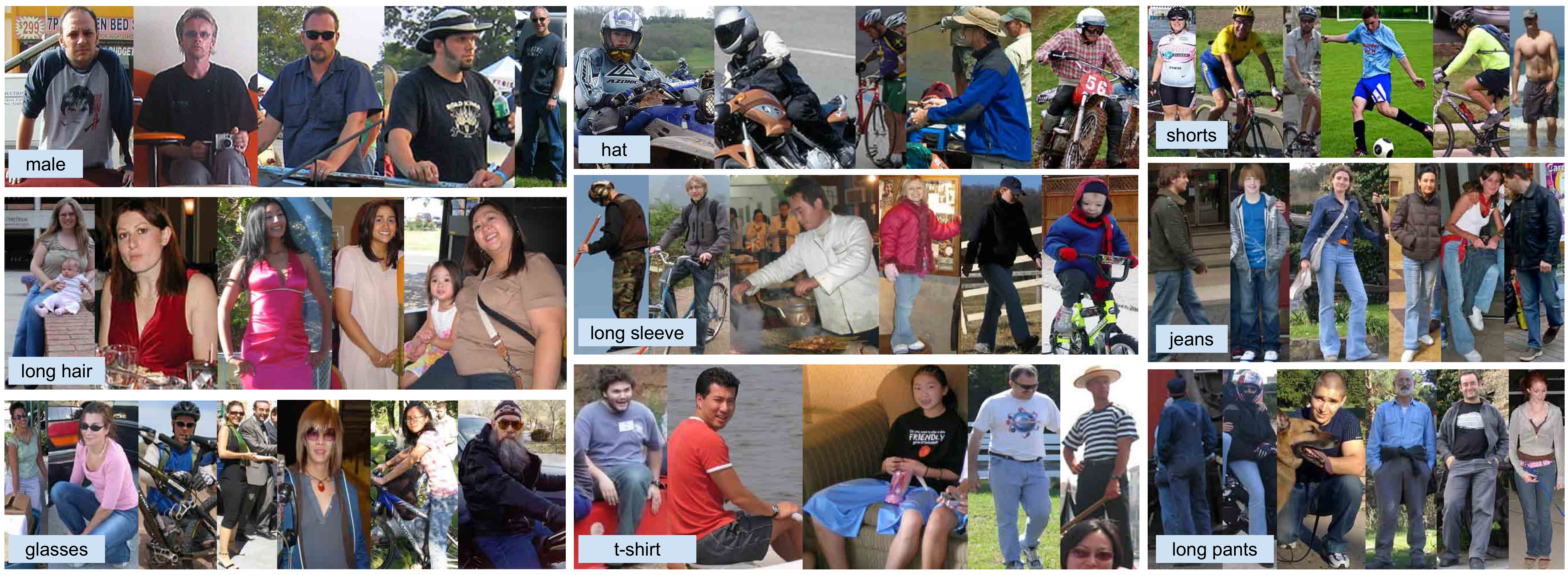}
		\vspace{-.5cm}
		\caption{Most positive attribute prediction on attributes of people dataset\cite{bourdev2011describing}. We cropped the image around the ground truth bounding box for display purpose.}
		\label{fig:outputpos}
		\vspace{-.5cm}
	\end{center}
\end{figure*}

\begin{figure*}[th!]
	\begin{center}
		\includegraphics[width=450px]{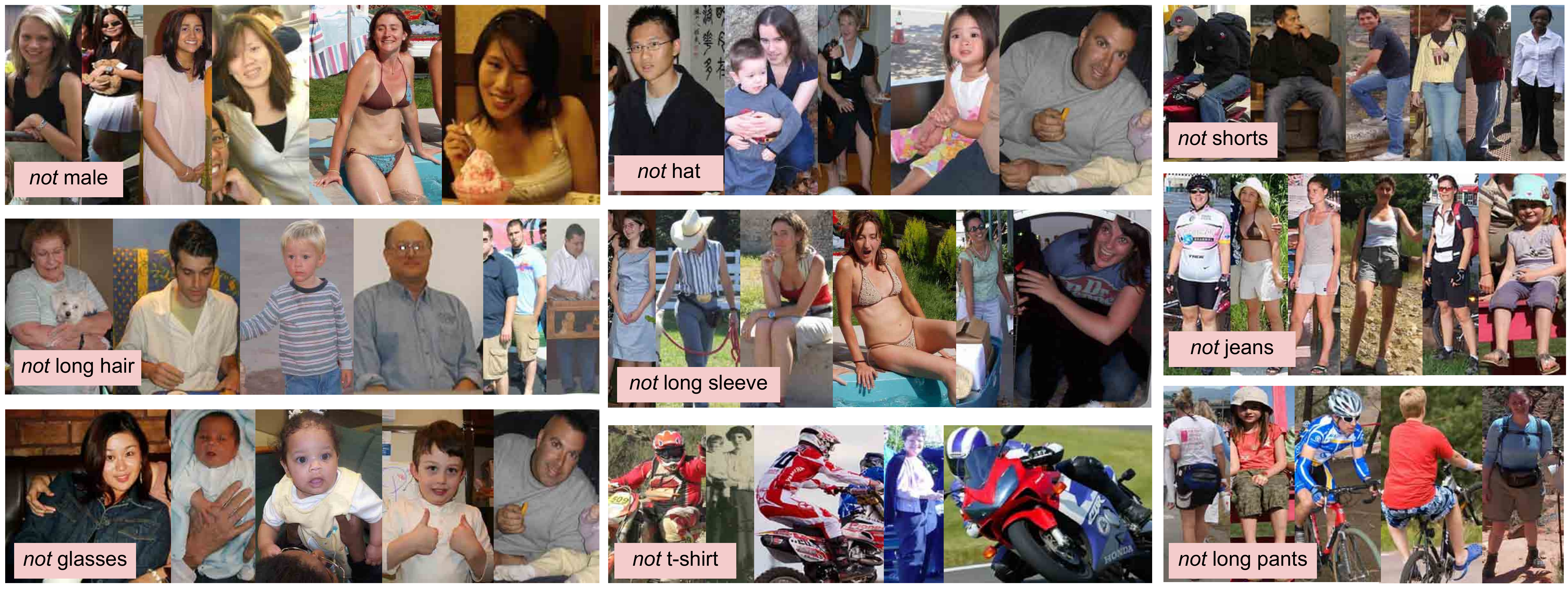}
		\vspace{-.5cm}
		\caption{Most negative attribute prediction on attributes of people dataset\cite{bourdev2011describing}. We cropped the image around the ground truth bounding box for display purpose.}
		\label{fig:outputneg}
		\vspace{-.5cm}
	\end{center}
\end{figure*}

\begin{figure*}[th!]
	\begin{center}
		\includegraphics[width=450px]{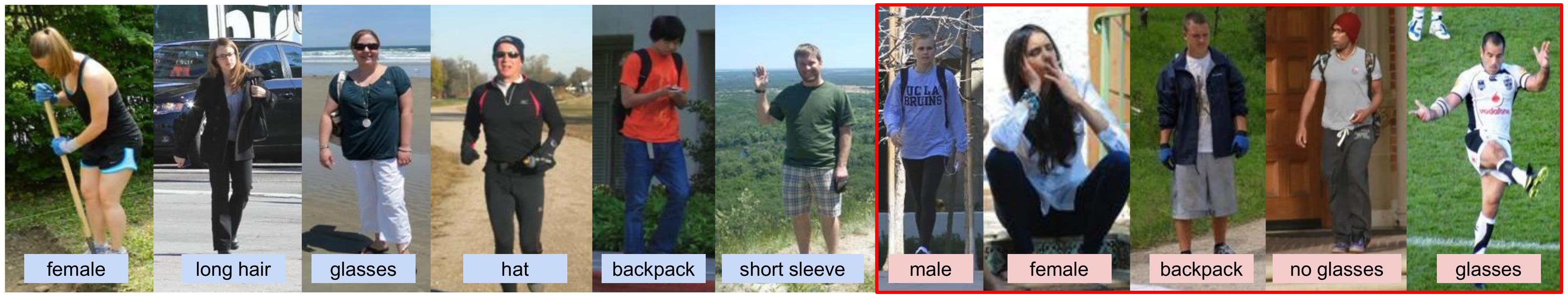}
		\vspace{-0.5cm}
		\caption{Output examples on \textbf{Pedestrian attribute dataset}. First six examples show successful examples with high scores. Rest of examples show failure examples with high scores. Actual testing images have large margin, but we crop the image around the ground truth bounding box for visualization.}
		\label{fig:result_ucla}
	\end{center}
\end{figure*}

\begin{figure*}[th!]
	\begin{center}
		\includegraphics[width=450pt]{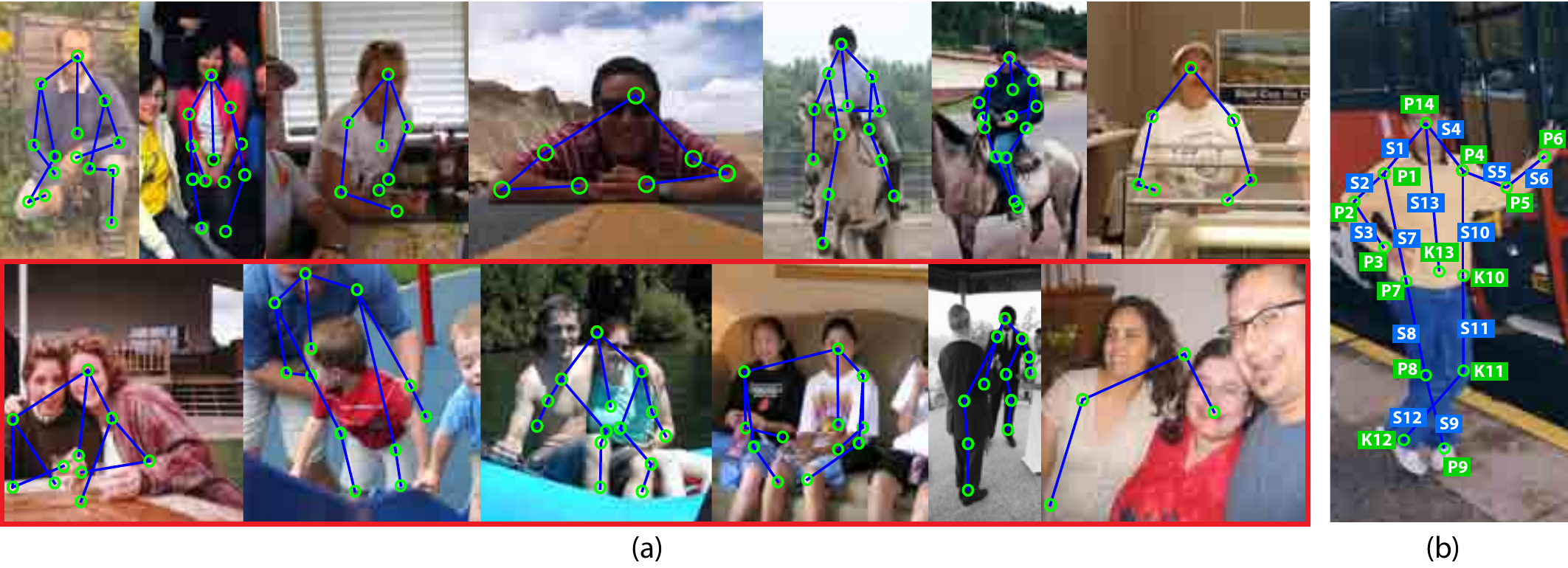}
		\vspace{-0.5cm}
		\caption{(a) Examples of pose-estimation result on Attributes of People dataset. First row shows success examples and second row shows failure examples. Most of failures are caused by multiple people in the image. (b) We define 14 keypoints and 13 sticks. }
		\label{fig:result_poselet_pose}
	\end{center}
\end{figure*}

\ifCLASSOPTIONcaptionsoff
\newpage
\fi


%

\ifCLASSOPTIONcompsoc
  \section*{Acknowledgments}
\else
  \section*{Acknowledgment}
\fi

This work was supported by a DARPA SIMPLEX project N66001-15-C-4035, ONR MURI project N00014-16-1-2007, and NSF IIS 1423305. The authors'd like to thank Dr. Brandon Rothrock for discussions.

\ifCLASSOPTIONcaptionsoff
  \newpage
\fi



\bibliographystyle{IEEEtran}
\bibliography{references}

\begin{thebibliography}{10}
\providecommand{\url}[1]{#1}
\csname url@samestyle\endcsname
\providecommand{\newblock}{\relax}
\providecommand{\bibinfo}[2]{#2}
\providecommand{\BIBentrySTDinterwordspacing}{\spaceskip=0pt\relax}
\providecommand{\BIBentryALTinterwordstretchfactor}{4}
\providecommand{\BIBentryALTinterwordspacing}{\spaceskip=\fontdimen2\font plus
\BIBentryALTinterwordstretchfactor\fontdimen3\font minus
  \fontdimen4\font\relax}
\providecommand{\BIBforeignlanguage}[2]{{%
\expandafter\ifx\csname l@#1\endcsname\relax
\typeout{** WARNING: IEEEtran.bst: No hyphenation pattern has been}%
\typeout{** loaded for the language `#1'. Using the pattern for}%
\typeout{** the default language instead.}%
\else
\language=\csname l@#1\endcsname
\fi
#2}}
\providecommand{\BIBdecl}{\relax}
\BIBdecl

\bibitem{farhadi2009describing}
A.~Farhadi, I.~Endres, D.~Hoiem, and D.~A. Forsyth, ``Describing objects by
  their attributes,'' in \emph{CVPR}, 2009.

\bibitem{bourdev2011describing}
L.~D. Bourdev, S.~Maji, and J.~Malik, ``Describing people: A poselet-based
  approach to attribute classification,'' in \emph{ICCV}, 2011.

\bibitem{patterson2012sun}
G.~Patterson and J.~Hays, ``Sun attribute database: discovering, annotating,
  and recognizing scene attributes,'' in \emph{CVPR}, 2012.

\bibitem{ramanan2006pose}
D.~Ramanan, ``Learning to parse images of articulated bodies,'' in \emph{NIPS},
  2006.

\bibitem{felzenszwalb2010object}
P.~F. Felzenszwalb, R.~B. Girshick, D.~McAllester, and D.~Ramanan, ``Object
  detection with discriminatively trained part-based models,'' in \emph{PAMI},
  vol.~32, no.~9.\hskip 1em plus 0.5em minus 0.4em\relax IEEE, 2010, pp.
  1627--1645.

\bibitem{rothrock2013integrating}
B.~Rothrock, S.~Park, and S.~C. Zhu, ``Integrating grammar and segmentation for
  human pose estimation,'' in \emph{CVPR}, 2013.

\bibitem{zhu2006stochastic}
S.~C. Zhu and D.~Mumford, ``A stochastic grammar of images,'' in
  \emph{Foundations and Trends in Computer Graphics and Vision}, vol.~2, no.~4,
  2006.

\bibitem{joo2013human}
J.~Joo, S.~Wang, and S.~C. Zhu, ``Human attribute recognition by rich
  appearance dictionary,'' in \emph{ICCV}, 2013.

\bibitem{zhang2013panda}
N.~Zhang, M.~Paluri, M.~Ranzato, T.~Darrell, and L.~Bourdev, ``Panda: Pose
  aligned networks for deep attribute modeling,'' in \emph{CVPR}, 2014.

\bibitem{zhang2013deformable}
N.~Zhang, R.~Farrell, F.~Iandola, and T.~Darrell, ``Deformable part descriptors
  for fine-grained recognition and attribute prediction,'' in \emph{ICCV},
  2013.

\bibitem{gkioxari2015actions}
G.~Gkioxari, R.~Girshick, and J.~Malik, ``Actions and attributes from wholes
  and parts,'' in \emph{ICCV}, 2015.

\bibitem{gkioxari2015rstarcnn}
G.~Gkioxari, R.~Girshick, and J.~MaliK, ``Contextual action recognition with
  r*cnn,'' in \emph{ICCV}, 2015.

\bibitem{chen2014articulated}
X.~Chen and A.~Yuille, ``Articulated pose estimation by a graphical model with
  image dependent pairwise relations,'' in \emph{NIPS}, 2014.

\bibitem{park2015attributed}
S.~Park and S.~C. Zhu, ``Attributed grammars for joint estimation of human
  attributes, part and pose,'' in \emph{ICCV}, 2015.

\bibitem{knuth1990attr}
D.~Knuth, ``The genesis of attribute grammars,'' in \emph{Proceedings of the
  International Conference on Attribute Grammars and Their Applications}, 1990.

\bibitem{abney1997stochastic}
S.~P. Abney, ``Stochastic attribute-value grammars,'' in \emph{Computational
  Linguistics}, vol.~23, no.~4.\hskip 1em plus 0.5em minus 0.4em\relax MIT
  Press, 1997, pp. 597--618.

\bibitem{han2009bottom}
F.~Han and S.~C. Zhu, ``Bottom-up/top-down image parsing with attribute
  grammar,'' in \emph{PAMI}, vol.~31, no.~1.\hskip 1em plus 0.5em minus
  0.4em\relax IEEE, 2009, pp. 59--73.

\bibitem{lin2009semantic}
L.~Lin, H.~Gong, L.~Li, and L.~Wang, ``Semantic event representation and
  recognition using syntactic attribute graph grammar,'' in \emph{Pattern
  Recognition Letters}, vol.~30, no.~2.\hskip 1em plus 0.5em minus 0.4em\relax
  Elsevier, 2009, pp. 180--186.

\bibitem{damen2012explaining}
D.~Damen and D.~Hogg, ``Explaining activities as consistent groups of events,''
  in \emph{IJCV}, vol.~98, no.~1.\hskip 1em plus 0.5em minus 0.4em\relax
  Springer, 2012, pp. 83--102.

\bibitem{wang2013weakly}
S.~Wang, J.~Joo, Y.~Wang, and S.~C. Zhu, ``Weakly supervised learning for
  attribute localization in outdoor scenes,'' in \emph{CVPR}, 2013.

\bibitem{liu2014single}
X.~Liu, Y.~Zhao, and S.~C. Zhu, ``Single-view 3d scene parsing by attributed
  grammar,'' in \emph{CVPR}, 2014.

\bibitem{cottrell1990empath}
G.~W. Cottrell and J.~Metcalfe, ``Empath: Face, emotion, and gender recognition
  using holons,'' in \emph{NIPS}, 1990.

\bibitem{golomb1991sexnet}
B.~A. Golomb, D.~Lawrence, and T.~Sejnowski, ``Sexnet: A neural network
  identifies sex from human faces,'' in \emph{NIPS}, 1991.

\bibitem{moghaddam2002learning}
B.~Moghaddam and M.-H. Yang, ``Learning gender with support faces,'' in
  \emph{PAMI}, 2002.

\bibitem{kwon1999age}
Y.~H. Kwon and N.~da~Vitoria~Lobo, ``Age classification from facial images,''
  in \emph{CVIU}, 1999.

\bibitem{kumar2011describable}
N.~Kumar, A.~C. Berg, P.~N. Belhumeur, and S.~K. Nayar, ``Describable visual
  attributes for face verification and image search,'' in \emph{PAMI}, vol.~33,
  2011.

\bibitem{bourdev2009poselets}
L.~D. Bourdev and J.~Malik, ``Poselets: Body part detectors trained using 3d
  human pose annotations,'' in \emph{ICCV}, 2009.

\bibitem{chen2012describing}
H.~Chen, A.~Gallagher, and B.~Girod, ``Describing clothing by semantic
  attributes,'' in \emph{ECCV}, 2012.

\bibitem{yang2011articulated}
Y.~Yang and D.~Ramanan, ``Articulated pose estimation with flexible
  mixtures-of-parts,'' in \emph{CVPR}, 2011.

\bibitem{fishler1973therepresentation}
M.~Fischler and R.~Elschlager, ``The representation and matching of pictorial
  structures,'' in \emph{IEEE Trans. on Computer}, 1973.

\bibitem{andriluka2009pictorial}
M.~Andriluka, S.~Roth, and B.~Schiele, ``Pictorial structures revisited: People
  detection and articulated pose estimation,'' in \emph{CVPR}, 2009.

\bibitem{eichner2009better}
M.~Eichner and V.~Ferrari, ``Better appearance models for pictorial
  structures,'' in \emph{BMVC}, 2009.

\bibitem{felzenszwalb2005pictorial}
P.~F. Felzenszwalb and D.~P. Huttenlocher, ``Pictorial structures for object
  recognition,'' in \emph{IJCV}, 2005.

\bibitem{sapp2010adaptive}
B.~Sapp, C.~Jordan, and B.~Taskar, ``Adaptive pose priors for pictorial
  structures,'' in \emph{CVPR}, 2010.

\bibitem{pedro2010cascade}
P.~F. Felzenszwalb, R.~B. Girshick, and D.~McAllester, ``Cascade object
  detection with deformable part models,'' in \emph{CVPR}, 2010.

\bibitem{girshick2011object}
R.~B. Girshick, P.~F. Felzenszwalb, and D.~McAllester, ``Object detection with
  grammar models,'' in \emph{NIPS}, 2011.

\bibitem{yang2013articulated}
Y.~Yang and D.~Ramanan, ``Articulated human detection with flexible mixtures of
  parts,'' in \emph{PAMI}, vol.~35, 2013.

\bibitem{pishchulin2013strong}
L.~Pishchulin, M.~Andriluka, P.~Gehler, and B.~Schiele, ``Strong appearance and
  expressive spatial models for human pose estimation,'' in \emph{ICCV}, 2013.

\bibitem{gkioxari2014kposelets}
G.~Gkioxari, B.~Hariharan, R.~Girshick, and J.~Malik, ``Using k-poselets for
  detecting people and localizing their keypoints,'' in \emph{CVPR}, 2014.

\bibitem{girshick2014rich}
R.~Girshick, J.~Donahue, T.~Darrell, and J.~Malik, ``Rich feature hierarchies
  for accurate object detection and semantic segmentation.'' in \emph{CVPR},
  2014.

\bibitem{johnson2011learning}
S.~Johnson and M.~Everingham, ``Learning effective human pose estimation from
  inaccurate annotation,'' in \emph{CVPR}, 2011.

\bibitem{Tompson2014Joint}
J.~J. Tompson, A.~Jain, Y.~LeCun, and C.~Bregler, ``Joint training of a
  convolutional network and a graphical model for human pose estimation,'' in
  \emph{NIPS}, 2014.

\bibitem{toshev2014deeppose}
A.~Toshev and C.~Szegedy, ``Deeppose: Human pose estimation via deep neural
  networks,'' in \emph{CVPR}, 2014.

\bibitem{fu1982syntactic}
K.~Fu, ``Syntactic pattern recognition and applicationsl,'' 1982.

\bibitem{geman2002composition}
S.~Geman, D.~Potter, and Z.~Chi, ``Composition systems,'' in \emph{Quarterly of
  Applied Mathematics}, no.~60, 2002, pp. 707--736.

\bibitem{fidler2006hierarchical}
S.~Fidler, G.~Berginc, and A.~Leonardis, ``Hierarchical statistical learning of
  generic parts of object structure,'' in \emph{CVPR}, 2006.

\bibitem{ren2015fasterrcnn}
S.~Ren, K.~He, R.~Girshick, and J.~Sun, ``Faster {R-CNN}: Towards real-time
  object detection with region proposal networks,'' in \emph{NIPS}, 2015.

\bibitem{girshick2015fastrcnn}
R.~Girshick, ``Fast r-cnn,'' in \emph{ICCV}, 2015.

\end{thebibliography}
%

%
\vspace{-1.5cm}
\begin{IEEEbiography}[{\includegraphics[width=1in,height=1.25in,clip,keepaspectratio]{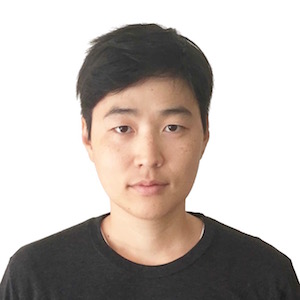}}]{Seyoung Park}
	received his M.S. degree in Computer Science from University of California, Los Angeles in 2012. He is now a Ph.D student at UCLA and a member of the  Center for Vision, Cognition, Learning, and Autonomy (VCLA). His research interests include pattern recognition, machine learning and computer vision, with a focus on attribute detection and recognition, and object detection with the statistical learning of hierarchical and compositional representations.
\end{IEEEbiography}

\vspace{-1.5cm}

\begin{IEEEbiography}[{\includegraphics[width=1in,height=1.25in,clip,keepaspectratio]{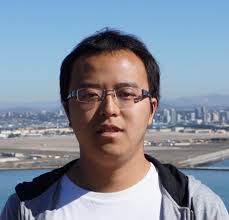}}]{Bruce X. Nie} received his B.S. degree in Computer Science from Zhengzhou University, Zhengzhou, China, in 2009. He worked as a Research Assistant during 2010 to 2011 at Lotus Hill Institute, Ezhou, China, and received his M.S. degree in Computer Science from Beijing Institute of Technology, Beijing, China, in 2012. Now He is a Ph.D student in the Department of Statistics at UCLA. His research interests include video surveillance, action detection and recognition and object detection.
\end{IEEEbiography}

\vspace{-1.0cm}

\begin{IEEEbiography}[{\includegraphics[width=1in,height=1.25in,clip,keepaspectratio]{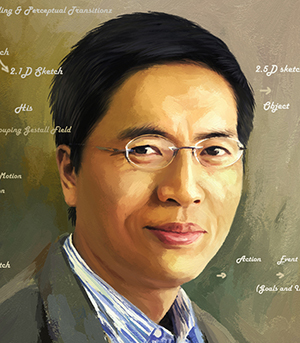}}]{Song-Chun Zhu}
	received his Ph.D. degree from Harvard University in 1996. He is currently professor of Statistics and Computer Science at UCLA, and  director of the  Center for Vision, Cognition, Learning and Autonomy (VCLA). He has published over 200 papers in computer vision, statistical modeling, learning, cognition, and visual arts. In recent years, his interest has also extended to cognitive robotics, robot autonomy, situated dialogues, and commonsense reasoning.  He received a number of honors, including the Helmholtz Test-of-time award in ICCV 2013, the Aggarwal prize from the Int'l Association of Pattern Recognition in 2008, the David Marr Prize in 2003 with Z. Tu et al., twice Marr Prize honorary nominations with Y. Wu et al. in 1999 for texture modeling and 2007 for object modeling respectively.  He received the Sloan Fellowship in 2001, a US NSF Career Award in 2001, and an US ONR Young Investigator Award in 2001. He is a Fellow of IEEE since 2011, and served as general co-chair for CVPR 2012.
\end{IEEEbiography}




\end{document}